\documentclass[conference]{IEEEtran}
\IEEEoverridecommandlockouts

\usepackage{cite}
\usepackage{amsmath,amssymb,amsfonts}
\usepackage{algorithm}
\usepackage{algorithmic}
\usepackage{graphicx}
\usepackage{textcomp}
\usepackage{xcolor}
\usepackage{subcaption}
\usepackage{booktabs}
\usepackage{verbatim}
\usepackage{comment}

\usepackage{booktabs}
\usepackage{graphicx}
\usepackage{makecell}

\usepackage{booktabs}
\usepackage{tabularx}
\usepackage{array}
\def\BibTeX{{\rm B\kern-.05em{\sc i\kern-.025em b}\kern-.08em
    T\kern-.1667em\lower.7ex\hbox{E}\kern-.125emX}}
\begin{document}

\title{HantaWatch: Federated Learning for Hantavirus Genomic Surveillance
}

\author{Shanika Iroshi Nanayakkara and Shiva Raj Pokhrel
\thanks{Authors are with the School of IT, Deakin University, Geelong, Australia; \textit{email: }{s222112938@deakin.edu.au, shiva.pokhrel@deakin.edu.au}}
}
\maketitle

\begin{abstract}
Hantavirus genomic surveillance is limited by distributed sequence data, non-IID source heterogeneity, and constrained expert-review capacity. We propose HantaWatch, a federated learning framework that enables laboratories and surveillance sites to collaboratively train sequence-based models without sharing raw data. HantaWatch integrates k-mer feature extraction, source-aware federated client construction, adaptive DU-FedProx optimization, surveillance-specific model selection, and prediction-only triage. Experiments on binary and multiclass tasks show that HantaWatch supports high-risk screening, outbreak-associated prediction, clade classification, and clinical-syndrome categorization while balancing predictive performance, false-negative risk, and update stability. The framework converts model output into risk scores, confidence estimates, uncertainty flags, and ranked expert-review priorities. HantaWatch therefore provides a practical federated decision-support layer for decentralized Hantavirus surveillance, supporting expert prioritization without replacing laboratory or public-health interpretation.
\end{abstract}


\section{Introduction}
\label{sec:introduction}
Hantaviruses~\cite{jonsson2010global} are rodent-borne zoonotic viruses that can cause severe human diseases, including hantavirus pulmonary syndrome, hantavirus cardiopulmonary syndrome, and hemorrhagic fever with renal syndrome. Although infections are less frequent than many respiratory diseases, they constitute a low-frequency, high-impact public-health threat~\cite{bennett2014reconstructing}. Surveillance remains difficult because cases are geographically dispersed, linked to ecological reservoirs, and shaped by diverse viral lineages, hosts, and regional transmission patterns. Early detection is further complicated by non-specific symptoms that resemble other febrile or respiratory illnesses. Therefore, timely clinical suspicion, laboratory confirmation, supportive care, and coordinated public-health response remain essential~\cite{world2023global}.

\begin{figure}[t]
\centering
\includegraphics[width=1\linewidth]{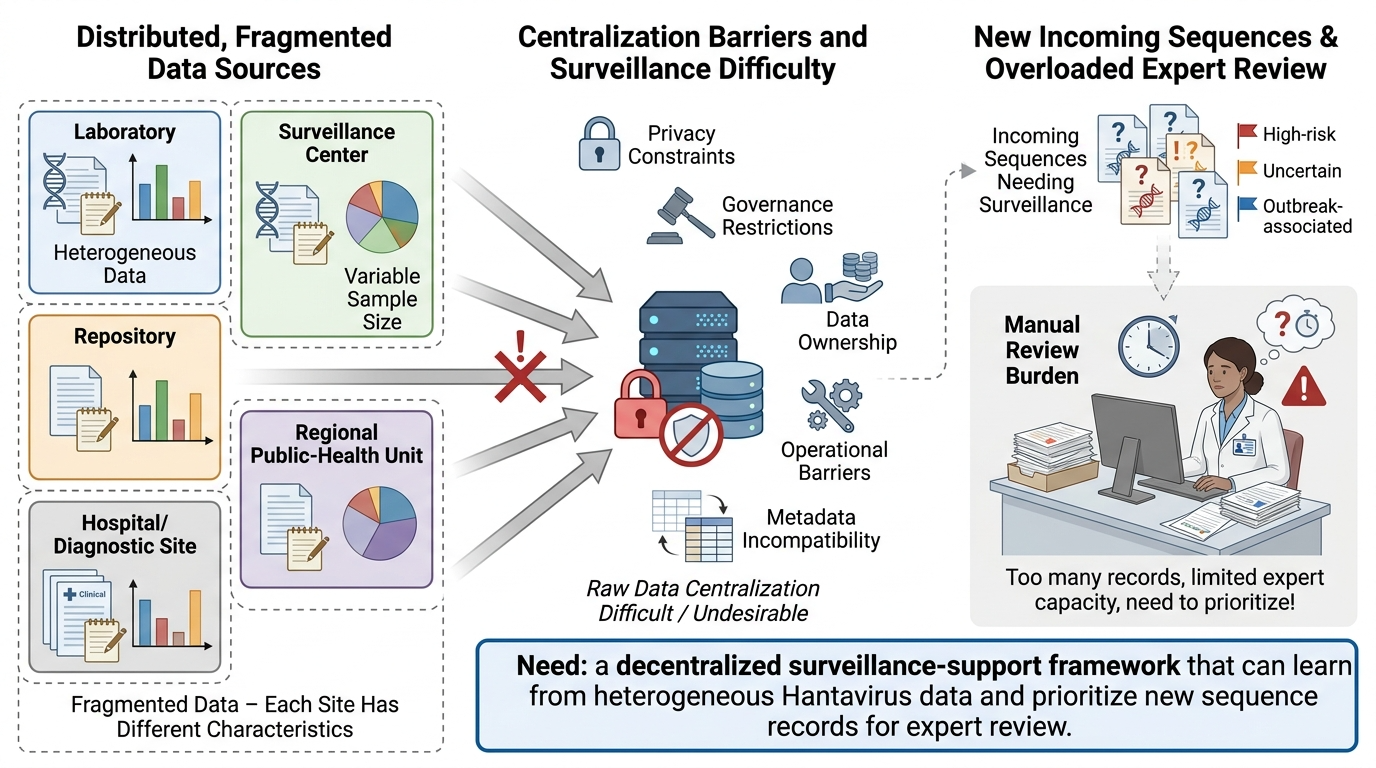}
\caption{\textit{Hantavirus FL problem.} Surveillance metadata and label distributions causing client drift, unstable updates, aggregation mismatch, and increased false-negative under standard aggregation.}
\label{fig:hantawatch_problem}
\end{figure}

Genomic and molecular surveillance enables sequence-level screening, pathogen classification, lineage monitoring, and metadata-informed risk assessment. Public repositories and curated hantavirus genome datasets provide valuable resources for computational surveillance modelling~\cite{kim2021genomic}. Sequence-derived representations, particularly $k$-mer features, allow machine-learning models to capture discriminative nucleotide patterns without full manual alignment~\cite{zhang2026novelkmer,moeckel2024surveykmer}. These models can support high-pathogenicity screening, outbreak-associated prediction, host-associated classification, clade classification, and clinical-syndrome categorization.

However, operational surveillance requires more than sequence classification. New records may arrive from multiple laboratories, repositories, regions, and surveillance sites, while expert-review capacity remains limited. The key operational question is not only what a model predicts, but which records should be reviewed first and why. High-risk-like, outbreak-associated-like, clinically relevant, low-confidence, or atypical sequences may require expedited expert inspection. An effective surveillance system should therefore produce predicted labels together with risk scores, confidence estimates, uncertainty indicators, review rationales, and priority rankings.

Existing genomic surveillance workflows provide essential capabilities for sequence retrieval, similarity search, phylogenetic interpretation, classification, visualization, and outbreak investigation~\cite{kim2021genomic}. Nevertheless, they do not fully address federated, adaptive, and uncertainty-aware prioritization of newly submitted hantavirus records under distributed and non-IID surveillance conditions. In practice, raw sequences and metadata are often held by different laboratories, agencies, repositories, or institutions, where centralization may be constrained by governance, privacy, operational, and structural limitations. Moreover, clients may differ in host distributions, clade composition, genome segments, outbreak histories, metadata completeness, and sampling bias, naturally creating a non-IID federated-learning environment.

Federated learning (FL)~\cite{11506081} is well suited to this setting because it enables collaborative model training while keeping raw data local~\cite{kairouz2021advances, 11458676}. FedAvg~\cite{mcmahan2017communication} provides a standard model-averaged baseline FL, but heterogeneous clients can generate conflicting local updates when their data distributions differ. FedProx~\cite{kairouz2021advances} partially mitigates this issue through proximal regularization, yet fixed learning rates, proximal coefficients, and local epoch schedules may remain inadequate across changing surveillance clients and rounds. In risk-sensitive genomic surveillance, such instability is critical: a model may achieve high accuracy while missing rare but important surveillance-positive records. 

Figure~\ref{fig:hantawatch_problem} illustrates the central challenge addressed in this work. Heterogeneous surveillance clients induce client drift and unstable local optimization, reducing the reliability of standard aggregation methods. A practical hantavirus surveillance framework must therefore solve two coupled problems: robust federated learning under non-IID sequence distributions and reliable decision support for expert-review prioritization.

To address these challenges, we introduce \textit{HantaWatch}, a federated genomic surveillance framework for hantavirus sequence-based screening, model suitability assessment, and expert-review prioritization. HantaWatch supports task-specific model training from distributed sequence records, comparison of candidate federated strategies, suitability-based model selection, and prediction-only analysis of new sequence batches. It is not an autonomous diagnostic or pathogenicity-determination system; rather, it is a decision-support layer that prioritizes records for laboratory, molecular, phylogenetic, clinical, or epidemiological review.

The main contributions of this work are as follows:

\begin{itemize}
\item We formulate Hantavirus genomic surveillance as a decentralized federated learning problem for expert-review prioritization without centralizing raw sequence data.

\item We introduce \textit{HantaWatch}, a federated surveillance-support framework that produces predicted labels, risk scores, uncertainty flags, review rationales, and ranked expert-review priorities.

\item We design source-aware non-IID client construction to capture realistic heterogeneity across hosts, clades, genome segments, outbreak contexts, clinical labels, metadata quality, and data sources.

\item We propose an adaptive Deep-Unfolded FedProx controller that maps client states to local learning rates, proximal coefficients, and epoch schedules to reduce client drift.

\item We develop a surveillance-specific model selection layer based on recall, false-negative rate, F1-score, AUC, macro-averaged metrics, and update stability rather than accuracy alone.

\end{itemize}







 We evaluate HantaWatch across binary and multiclass Hantavirus surveillance tasks, including high-pathogenicity screening, outbreak-associated prediction, clade classification, and clinical-syndrome categorization, under both controlled non-IID and source-representative federated settings. We demonstrate that HantaWatch can translate trained federated models into prediction-only expert-review workflows, thereby supporting prioritization of high-risk, clinically relevant, outbreak-associated, or uncertain records while preserving the role of human experts in final interpretation.

 The core methodological component of HantaWatch is a Deep-Unfolded FedProx controller for adaptive federated optimization. Instead of using fixed local training settings, HantaWatch unfolds the federated process across communication rounds and maps client-level states---including previous loss, gradient norm, update norm, class imbalance, and round index---to adaptive controls. These controls determine the local learning rate $\eta$, proximal coefficient $\mu$, and number of local epochs $E$, enabling each client to adjust its training behaviour according to local heterogeneity and update stability.

HantaWatch also treats the local optimizer as a configurable component of the surveillance workflow. Candidate federated methods are evaluated across optimizer settings such as SGD, Adam, AdamW, and RMSProp, allowing the framework to assess convergence, predictive performance, false-negative risk, and update stability. This avoids attributing performance gains to a single favourable optimizer configuration and strengthens method-level evaluation.

\section{Related Work}
\label{sec:related_work}

Genomic surveillance has become central to pathogen monitoring because it enables sequence-level classification, lineage tracking, outbreak investigation, and integration of molecular and epidemiological evidence~\cite{getchell2024pathogen}. Sequencing-based surveillance complements conventional epidemiology by providing higher-resolution insight into pathogen diversity, transmission dynamics, and emerging variants~\cite{tiwari2025genomics,getchell2024pathogen,engelthaler2024genomic}. For Hantaviruses, genomic and phylogenetic analysis is particularly important because viral diversity is shaped by reservoir hosts, geography, genome segments, and regional circulation patterns. Recent studies of Hantaan virus further demonstrate the value of genomic surveillance to characterize orthohantavirus diversity and inform public-health response~\cite{park2025epidemiological,noh2025phylogenetic}.

Several genomic surveillance platforms provide important capabilities for pathogen analysis. HantaNet, implemented within MicrobeTrace, supports curated hantavirus reference data, classification, genomic epidemiology, phylogenetic visualization, mapping, network analysis, and dashboards~\cite{cintron2023hantanet,campbell2021microbetrace,shankar2026microbetrace}. Broader platforms such as Nextstrain, Pathogenwatch, and NCBI Virus support real-time pathogen evolution analysis, genome upload and interpretation, relatedness assessment, typing or lineage analysis, and large-scale viral sequence retrieval~\cite{hadfield2018nextstrain,argimon2021rapid,alikhan2026pathogenwatch,goldfarb2025ncbi,brister2015ncbi}. Similarly, tools and initiatives such as Nextclade, Pangolin, UShER, COG-UK, CDC SPHERES, WHO genomic surveillance frameworks, and CoV-Spectrum support quality assessment, clade or lineage assignment, phylogenetic placement, sequencing coordination, metadata integration, and variant monitoring~\cite{turakhia2021ultrafast,covid2020integratedCOG,centers2022sars,world2023global,chen10analysis}.



Despite these advances, existing systems mainly focus on centralized or uploaded sequence analysis, phylogenetic interpretation, lineage assignment, visualization, and public-health coordination. They do not directly address federated learning from distributed Hantavirus sequence holders, where raw data may remain under local institutional, governance, or operational control. Moreover, they do not provide an adaptive non-IID learning mechanism that links source-heterogeneous model training with uncertainty-aware prioritization of newly submitted records for expert review.

\textit{HantaWatch} addresses this gap by complementing existing genomic surveillance infrastructure with a federated expert-review prioritization layer. Unlike HantaNet, MicrobeTrace, Nextstrain, Pathogenwatch, and related platforms, HantaWatch is designed to support distributed model training, source-aware non-IID adaptation, surveillance-specific model suitability assessment, uncertainty-aware triage, and ranked expert-review outputs for new Hantavirus sequences.

\begin{figure*}[th]
    \centering
    \includegraphics[width=1\linewidth]{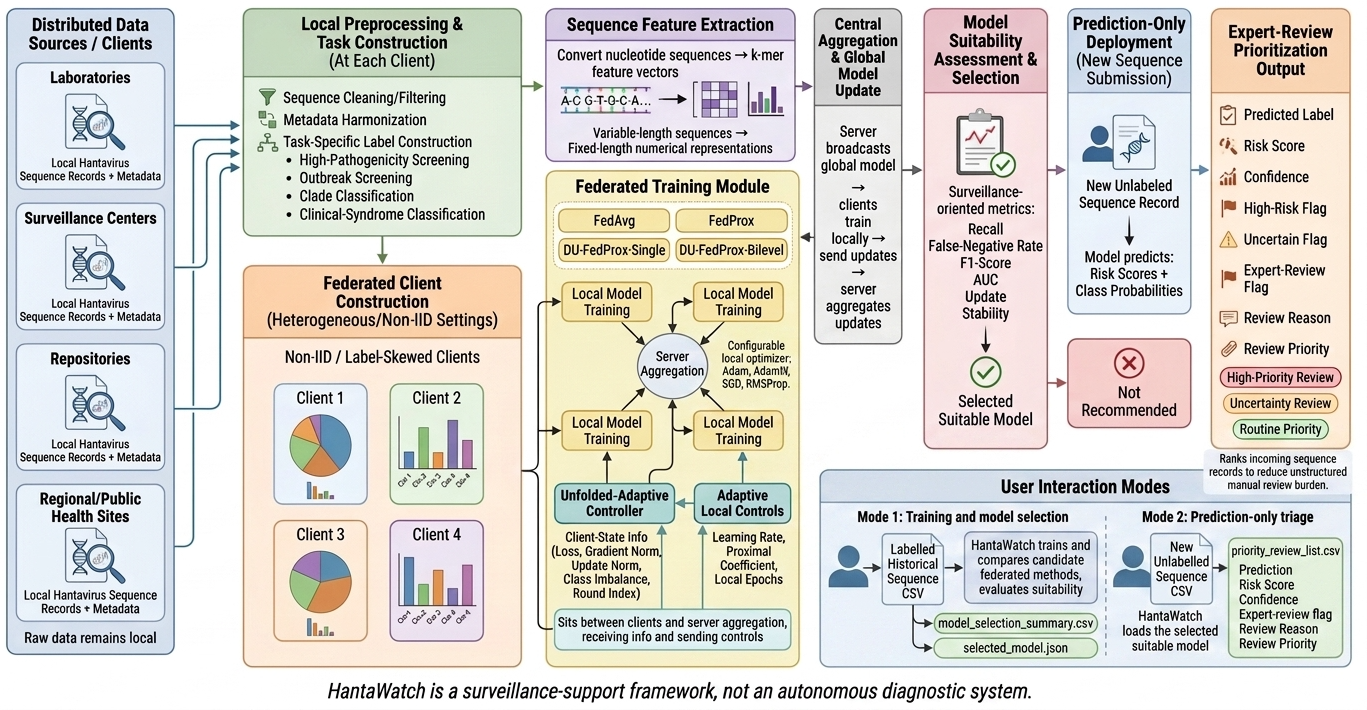}
    \caption{HantaWatch: FL Framework for Hantavirus Surveillance and Expert-Review Prioritization}
    \label{fig:framework}
\end{figure*}


\section{Proposed HantaWatch Methodology}
\label{sec:methodology}

\subsection{Uncertainty-Aware Decision Support and Expert Triage}

In high-consequence biomedical and public-health contexts, model predictions should not be regarded as autonomous decisions. The importance of uncertainty estimation, out-of-distribution detection, and decision-support design is increasingly acknowledged for establishing trustworthy machine learning. In bioinformatics, out-of-distribution learning is especially pertinent because genomic sequences can deviate from the training distribution due to rare lineages, incomplete sampling, sequencing artifacts, host shifts, or limited reference coverage. In clinical and biomedical artificial intelligence, uncertainty-aware systems are typically conceptualized as tools that support, rather than replace, expert decision-making.

This perspective is directly applicable to HantaWatch. A high-risk prediction should not be considered definitive evidence of pathogenicity, nor should a low-confidence prediction be disregarded. Instead, HantaWatch converts predictions into expert-review priorities. Records identified as high-risk-like, outbreak-associated-like, clinically relevant, or uncertain are flagged for expedited expert review, whereas routine records are assigned to standard surveillance workflows. The framework supports the allocation of expert workload by generating risk scores, confidence values, uncertainty flags, review rationales, and priority outputs ranked.

\subsection{Positioning of HantaWatch}

The proposed framework operates at the intersection of genomic surveillance platforms, viral sequence machine learning, federated biomedical learning, non-independent and identically distributed (non-IID) optimization, and uncertainty-awarekey aspects Existing systems address key aspects of pathogen surveillance, such as sequence retrieval, phylogenetic analysis, public health, classification, and public-health reporting. However, to our knowledge, current Hantavirus and pathogen surveillance tools do not offer the specific combination of federated non-IID learning, DU-FedProx adaptive optimization, model suitability assessment, uncertainty-aware review prioritization, and user-facing new-sequence triage. HantaWatch addresses this gap by transforming distributed Hantavirus sequence learning into an operational expert-review prioritization workflow.

Existing tools facilitate the organization, filtering, classification, and visualization of genomic sequences. HantaWatch introduces a triage layer that transforms unlabeled hantavirus sequences into a ranked list for expert review, prioritized by predicted surveillance relevance and associated uncertainty.

\subsection{System Overview}
\label{subsec:system_overview}

Let $\mathcal{D}_m$ represent the local dataset for client $m \in \{1,\ldots,M\}$, where each client may be a laboratory, regional surveillance center, repository, or public-health site. Each dataset includes Hantavirus nucleotide sequence records and, when available, metadata such as host, segment, clade, outbreak status, clinical syndrome, source, or collection details. The server manages training by broadcasting global model parameters and aggregating local model updates, ensuring that raw client sequences and metadata remain local.

The HantaWatch workflow comprises seven primary stages:
\begin{enumerate}
    \item Harmonization of local data and task construction;
    \item sequence feature extraction using $k$-mer representations;
    \item federated client construction under heterogeneous non-IID settings;
    \item federated training using candidate methods;
    \item adaptive local control using DU-FedProx;
    \item model suitability assessment and model selection;
    \item prediction-only triage and expert-review prioritization.
\end{enumerate}

The framework is not intended to replace virological, laboratory, or epidemiological assessments. Rather, HantaWatch serves as a surveillance-support layer to identify records that may require accelerated expert review based on learned sequence patterns, prediction confidence, and task-specific risk indicators.

\subsection{Data Harmonization and Surveillance Task Construction}
\label{subsec:data_harmonization}

Each input dataset is harmonized into a standardized task-specific format. At a minimum, every supervised record includes a sample identifier, nucleotide sequence, and task label. Optional metadata fields may include host, host category, segment, clade, clinical syndrome, outbreak association, source, and collection information. Records lacking sequences, containing invalid nucleotide content, or missing task labels are excluded from supervised training for the relevant task.

Given a sequence record $s_i$, task label $y_i$, and optional metadata vector $\mathbf{z}_i$, the task-specific dataset is represented as
\begin{equation}
    \mathcal{D}
    =
    \{(s_i, y_i, \mathbf{z}_i)\}_{i=1}^{N},
\end{equation}
where $N$ is the number of usable labeled records for the selected task. The label $y_i$ depends on the surveillance task. For binary tasks, labels are encoded as
\begin{equation}
    y_i \in \{0,1\},
\end{equation}
where class $1$ denotes the surveillance-positive condition. For multiclass tasks, such as clade classification and clinical-syndrome classification, labels are encoded as
\begin{equation}
    y_i \in \{0,1,\ldots,K-1\},
\end{equation}
where $K$ is the number of task-specific classes.


\subsection{Sequence Feature Extraction}
\label{subsec:feature_extraction}

Variable-length nucleotide sequences are transformed into fixed-length numerical representations using $k$-mer feature extraction \cite{zhang2026novelkmer}. A $k$-mer refers to a contiguous subsequence of length $k$ extracted from a nucleotide sequence. For each sequence $s_i$, the feature extraction function $\Phi_k(\cdot)$ maps the sequence into a fixed-dimensional vector:

\begin{equation}
    \mathbf{x}_i = \Phi_k(s_i) \in \mathbb{R}^{d},
\end{equation}
where $d$ is the number of retained $k$-mer features. In the main experiments, $k=4$ was used. The resulting vector $\mathbf{x}_i$ contains normalized $k$-mer counts or frequencies and provides an alignment-independent representation suitable for conventional machine-learning and federated-learning models.

After feature extraction, the supervised dataset for a given task becomes
\begin{equation}
    \mathcal{D}
    =
    \{(\mathbf{x}_i,y_i)\}_{i=1}^{N}.
\end{equation}
For multiclass tasks, labels are encoded as class indices and evaluated using macro-averaged metrics to account for class imbalance.

\subsection{Federated Client Construction}
\label{subsec:federated_client_construction}

The processed dataset is partitioned into $M$ federated clients:
\begin{equation}
    \mathcal{D}
    =
    \bigcup_{m=1}^{M} \mathcal{D}_m,
    \qquad
    \mathcal{D}_m \cap \mathcal{D}_{m'} = \emptyset
    \quad \text{for } m \neq m',
\end{equation}
where $\mathcal{D}_m = \{(\mathbf{x}_i,y_i)\}_{i=1}^{n_m}$ is the local dataset of client $m$, and $n_m = |\mathcal{D}_m|$. In this study, $M=4$ clients were used for controlled federated experiments.

To model realistic surveillance heterogeneity, we generate non-IID client distributions using label-skewed partitions. This approach mirrors practical Hantavirus surveillance, where laboratories, repositories, or regions may encounter varying host distributions, clade compositions, outbreak histories, sequence completeness, or clinical-syndrome labels. The empirical label distribution for client $m$ is defined as
\begin{equation}
    p_m(y)
    =
    \frac{1}{n_m}
    \sum_{i \in \mathcal{D}_m}
    \mathbb{I}(y_i = y),
\end{equation}
where $\mathbb{I}(\cdot)$ is the indicator function. Under non-IID splitting, $p_m(y)$ may differ substantially across clients and from the global label distribution. This setting is central to the evaluation because label skew can induce client drift, unstable local updates, and reduced global model reliability.

\subsection{Federated Evaluation Settings}
\label{subsec:federated_evaluation_settings}

HantaWatch was evaluated using two complementary federated settings to differentiate controlled heterogeneity analysis from source-level surveillance validation. The first setting employs simulated non-independent and identically distributed (non-IID) client construction from a processed dataset. The second setting designates independent data sources as federated clients. These approaches enable the framework to be assessed under both reproducible experimental partitions and realistic source-driven heterogeneity.

\subsubsection{Setting I: Controlled Non-IID Client Simulation}

Within the controlled federated setting, the processed dataset $\mathcal{D}$ is partitioned among $M$ clients using label-skewed non-independent and identically distributed (non-IID) splitting. This approach evaluates the stability of candidate federated methods when clients encounter varying label distributions. This scenario closely mirrors practical Hantavirus surveillance, in which client-specific data may vary by host type, clade composition, outbreak association, sequence completeness, geographic origin, or clinical label availability.

This experimental setting establishes a controlled benchmark, as all candidate methods are evaluated using an identical feature extraction pipeline, model architecture, client partition, training rounds, and evaluation protocol. Consequently, observed performance differences can be attributed primarily to the federated learning method and optimizer configuration, rather than to variations in data preparation. In this study, the controlled non-IID experiments utilized $M=4$ clients and were conducted for each candidate method, including FedAvg, FedProx, DU-FedProx-Single, and DU-FedProx-Bilevel.

This experimental setting is designed to evaluate robustness to artificial yet reproducible client heterogeneity. Specifically, it investigates whether adaptive proximal control mitigates the negative effects of label skew, including client drift, unstable local updates, reduced recall, increased false-negative rate, and poor global generalization.

\subsubsection{Setting II: Source-as-Client Federated Validation}

In the source-as-client framework, each independent data source or repository functions as a federated client. Rather than generating clients via artificial partitioning, client identity is defined by the provenance of the sequence data. Formally, if $\mathcal{S}={1,\ldots,S}$ denotes the set of available sources, the federated dataset is represented as
\begin{equation}
\mathcal{D}
=
\bigcup_{s=1}^{S} \mathcal{D}^{(s)},
\qquad
\mathcal{D}^{(s)} \cap \mathcal{D}^{(s')} = \emptyset
\quad \text{for } s \neq s',
\end{equation}
where $\mathcal{D}^{(s)}$ denotes the labeled sequence records obtained from source $s$. In this setting, each source acts as one federated client and retains its own source-specific label distribution.

This setting is significant because actual Hantavirus surveillance data are rarely derived from a single homogeneous population. Various repositories, laboratories, studies, and surveillance programs may encompass distinct host categories, genome segments, geographic regions, outbreak histories, sequencing completeness, and annotation protocols. Consequently, source-as-client validation offers a more realistic assessment of whether HantaWatch can function effectively across heterogeneous data sources without presuming centralized pooling or identical data distributions.

The source-as-client setting aims to assess both both classification performance and the continued suitability of the selected federated model for prediction-only deployment in the presence of client heterogeneity resulting from source differences. This approach aligns with the surveillance-oriented design of HantaWatch, which requires the identification of reliable models prior to generating expert-review priority lists for unlabelled sequences.

\subsubsection{Role of the Two Settings}

The two federated settings fulfill distinct, yet complementary, roles. The controlled non-IID setting offers a reproducible experimental benchmark for comparing federated methods under defined label-skew conditions. In contrast, the source-as-client setting assesses whether the same modeling approach remains effective under more realistic, source-driven heterogeneity. Employing both settings enhances the robustness of the evaluation, as methods that perform well solely under artificial partitions may not generalize to independent surveillance sources, while methods evaluated only on source-level clients may be challenging to analyze under controlled heterogeneity.


\subsection{DU-FedProx Adaptive Controller}
\label{subsec:dufedprox_controller}

\begin{figure}
    \centering
    \includegraphics[width=1\linewidth]{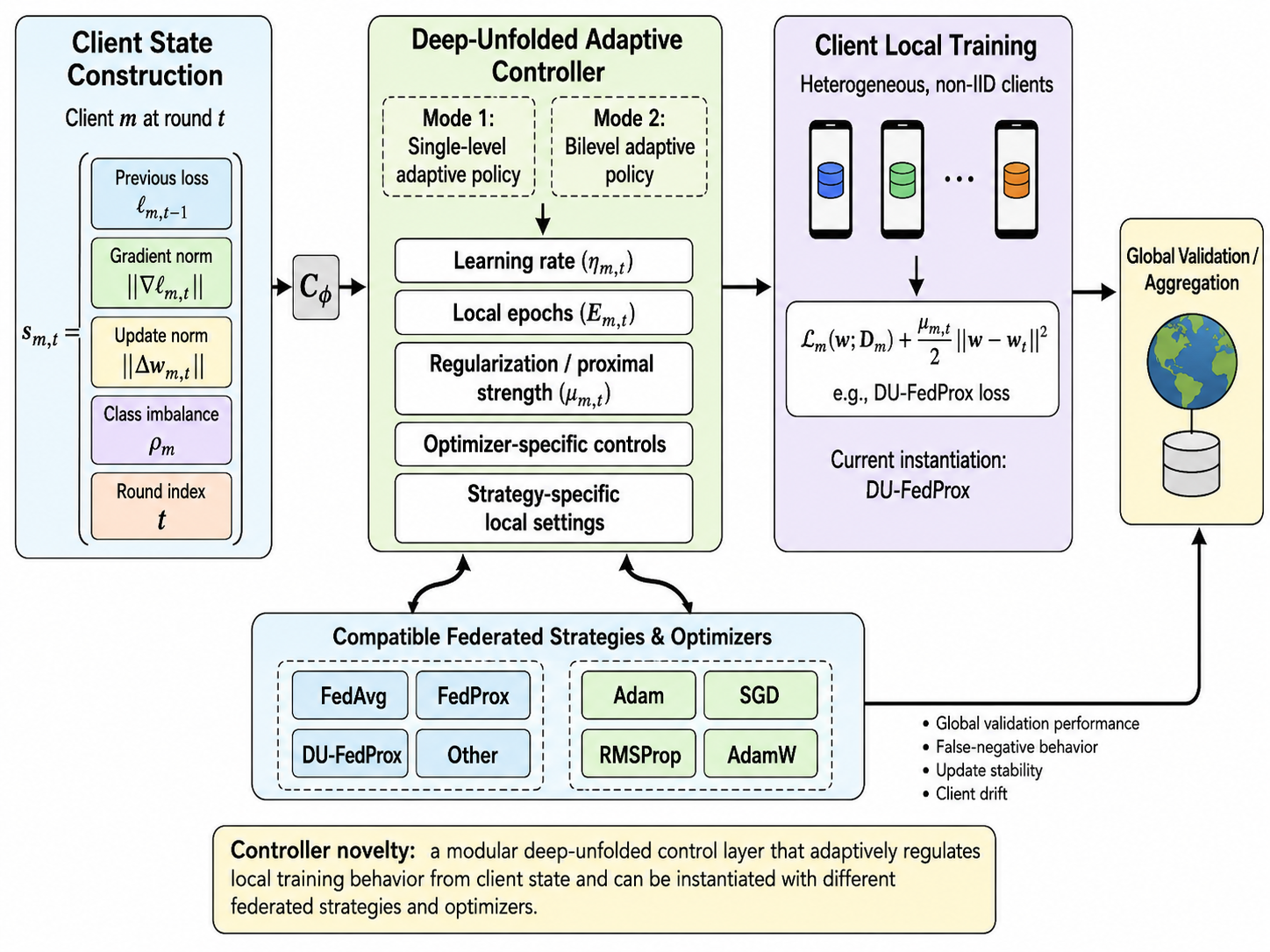}
    \caption{Adaptive Controller Process}
    \label{fig:placeholder}
\end{figure}

\begin{algorithm}[t]
\caption{DU-FedProx Local Training with Adaptive Controller}
\label{alg:dufedprox_training}
\begin{algorithmic}[1]
\REQUIRE Global model $\mathbf{w}_t$, client dataset $\mathcal{D}_m$, controller $\mathcal{C}_{\phi}$, round $t$
\ENSURE Updated client model $\mathbf{w}_{m,t}$ and diagnostics
\STATE Compute client state $\mathbf{s}_{m,t}$ using loss, gradient norm, update norm, class imbalance, and round index.
\STATE Generate adaptive controls:
\[
(\eta_{m,t}, \mu_{m,t}, E_{m,t})
=
\mathcal{C}_{\phi}(\mathbf{s}_{m,t}).
\]
\STATE Initialize local model $\mathbf{w}_{m,t}^{(0)} \leftarrow \mathbf{w}_t$.
\FOR{$e=1,\ldots,E_{m,t}$}
    \STATE Minimize the adaptive proximal objective:
    \[
    \mathcal{L}_{m}(\mathbf{w};\mathcal{D}_m)
    +
    \frac{\mu_{m,t}}{2}
    \left\|
    \mathbf{w}-\mathbf{w}_t
    \right\|_2^2.
    \]
    \STATE Apply a local optimizer step using learning rate $\eta_{m,t}$.
\ENDFOR
\STATE Compute local diagnostics: loss, gradient norm, update norm, and class-imbalance state.
\RETURN Updated local model $\mathbf{w}_{m,t}$ and diagnostics.
\end{algorithmic}
\end{algorithm}

While FedProx effectively reduces client drift, it generally employs fixed local training controls. In heterogeneous surveillance environments, clients may necessitate distinct learning rates, proximal strengths, or local epoch schedules based on local loss dynamics, gradient magnitude, update stability, and label imbalance. To overcome this limitation, HantaWatch implements a Deep-Unfolded FedProx controller.

At each communication round $t$, client $m$ constructs a state vector
\begin{equation}
    \mathbf{s}_{m,t}
    =
    \left[
    \ell_{m,t-1},
    \left\|\nabla \ell_{m,t}\right\|_2,
    \left\|\Delta \mathbf{w}_{m,t-1}\right\|_2,
    \rho_m,
    \frac{t}{T}
    \right]^{\top},
    \label{eq:client_state}
\end{equation}
where $\ell_{m,t-1}$ is the previous local loss, $\left\|\nabla \ell_{m,t}\right\|_2$ is the gradient norm, $\left\|\Delta \mathbf{w}_{m,t-1}\right\|_2$ is the previous update norm, $\rho_m$ is a class-imbalance score, and $t/T$ is the normalized round index. For binary tasks, $\rho_m$ can be computed from the positive-class rate. For multiclass tasks, $\rho_m$ can be computed from the deviation between the local class distribution and a uniform or global class distribution.

The controller maps the client state to adaptive local training controls:
\begin{equation}
    (\eta_{m,t}, \mu_{m,t}, E_{m,t})
    =
    \mathcal{C}_{\phi}(\mathbf{s}_{m,t}),
    \label{eq:controller_mapping}
\end{equation}
where $\eta_{m,t}$ is the local learning rate, $\mu_{m,t}$ is the adaptive proximal coefficient, $E_{m,t}$ is the number of local epochs, and $\mathcal{C}_{\phi}$ denotes the controller parameterized by $\phi$.

The DU-FedProx local objective is therefore
\begin{equation}
    \mathcal{L}_{m,t}^{\mathrm{DU}}(\mathbf{w})
    =
    \mathcal{L}_{m}(\mathbf{w};\mathcal{D}_m)
    +
    \frac{\mu_{m,t}}{2}
    \left\|
    \mathbf{w} - \mathbf{w}_t
    \right\|_2^2.
    \label{eq:dufedprox_objective}
\end{equation}

This study evaluates two types of controller. DU-FedProx-Single employs a single-level adaptive policy to generate $\eta_{m,t}$, $\mu_{m,t}$, and $E_{m,t}$ based on the current state of each client. In contrast, DU-FedProx-Bilevel updates the controller using an outer objective that considers post-aggregation surveillance performance, false negatives, update stability, and client drift. Consequently, the bilevel controller adapts to both local client behavior and the impact of local decisions on overall surveillance performance.

\subsection{Model Suitability Assessment and Selection}
\label{subsec:model_suitability}

A primary design principle of HantaWatch is that a trained model should not be deployed for surveillance solely based on high accuracy. In surveillance contexts where risk is a significant concern, missing high-risk or outbreak-related records poses a greater threat than generating some false alarms, as such omissions can delay expert review. Consequently, HantaWatch evaluates each method using criteria specifically tailored to surveillance requirements.

For binary tasks, evaluation metrics include accuracy, precision, recall, specificity, F1-score, area under the curve (AUC), false-negative rate (FNR), and update norm \cite{saito2015precision}. For multiclass tasks, macro-F1, macro-recall, macro-AUC, and macro-FNR are employed to ensure that suboptimal performance on minority classes is not obscured by high overall accuracy. Update norm is included to assess training stability, as large or unstable client updates may indicate client drift or aggregation issues.

Let $q(a)$ represent the suitability score for candidate method $a \in \mathcal{M}$. HantaWatch ranks candidate models according to task-specific metric priorities. For binary tasks with positive surveillance, the framework emphasizes recall, false negative rate (FNR), F1-score, area under the curve (AUC) and update stability. For multiclass tasks, macro-F1, macro-recall, macro-FNR, and update stability are prioritized. A candidate model is selected only if it meets the minimum suitability criteria; otherwise, the framework issues a not-recommended decision.
\begin{equation}
    a^{\star}
    =
    \arg\max_{a \in \mathcal{M}_{\mathrm{suitable}}}
    q(a),
\end{equation}
where $\mathcal{M}_{\mathrm{suitable}} \subseteq \mathcal{M}$ denotes the set of methods satisfying the task-specific suitability thresholds. If $\mathcal{M}_{\mathrm{suitable}} = \emptyset$, HantaWatch does not recommend prediction-only deployment.

At this stage, the result is either a suitable chosen model or a decision not to recommend any. This safety-focused approach stops weak or unstable models from being automatically used for expert review prioritization.

\subsection{Prediction-Only Triage and Expert-Review Prioritization}
\label{subsec:prediction_triage}

After selecting an appropriate model, HantaWatch can be used to analyze new unlabelled Hantavirus sequence records in prediction-only mode. These records undergo the same sequence cleaning and $k$-mer feature extraction procedures as those applied during training. The model subsequently generates predicted labels and class probabilities.

For a binary task, the model outputs a surveillance-positive probability
\begin{equation}
    r_i = P(y_i = 1 \mid \mathbf{x}_i),
\end{equation}
which is interpreted as a task-specific risk score. The prediction confidence is computed as
\begin{equation}
    c_i = \max_{y} P(y \mid \mathbf{x}_i).
\end{equation}
For multiclass tasks, the predicted class is
\begin{equation}
    \hat{y}_i = \arg\max_{y} P(y \mid \mathbf{x}_i).
\end{equation}

HantaWatch converts these predictions into outputs intended for expert review. Each new record receives a predicted label, risk score, confidence value, high-risk flag, uncertain flag, expert-review flag, review reason, and a review-priority category. Records are flagged for review if they exhibit a high risk score, low confidence, predictions near the decision boundary, high uncertainty, or labels relevant to surveillance tasks. In this study, the priority categories include high-priority review, uncertainty review, and routine priority.

The triage layer enables experts to manage their workload more effectively. Records identified as high-risk, outbreak-associated, clinically relevant, clade-relevant, or uncertain are prioritized for expedited review, whereas routine records are directed to standard surveillance processes. 

\begin{algorithm}[t]
\caption{HantaWatch Training and Model-Selection Workflow}
\label{alg:hantawatch_training}
\begin{algorithmic}[1]
\REQUIRE Labelled Hantavirus sequence datasets $\{\mathcal{D}_m\}_{m=1}^{M}$, candidate methods $\mathcal{M}$, number of rounds $T$
\ENSURE Selected suitable model or not-recommended decision
\STATE Harmonize sequence records, metadata fields, and task labels.
\STATE Extract $k$-mer features $\mathbf{x}_i = \Phi_k(s_i)$.
\STATE Construct binary or multiclass task labels.
\STATE Partition data into non-IID federated clients.
\FOR{each candidate method $a \in \mathcal{M}$}
    \STATE Initialize global model $\mathbf{w}_0$.
    \FOR{$t=0,1,\ldots,T-1$}
        \STATE Server broadcasts $\mathbf{w}_t$ to clients.
        \FOR{each client $m=1,\ldots,M$}
            \STATE Client trains locally using method $a$.
            \STATE Client returns $\mathbf{w}_{m,t}$ or $\Delta \mathbf{w}_{m,t}$ and diagnostics.
        \ENDFOR
        \STATE Server aggregates client models to obtain $\mathbf{w}_{t+1}$.
    \ENDFOR
    \STATE Evaluate final model using surveillance-oriented metrics.
\ENDFOR
\STATE Apply suitability criteria to all candidate methods.
\IF{at least one method is suitable}
    \STATE Select best suitable model $a^{\star}$.
\ELSE
    \STATE Return not-recommended decision.
\ENDIF
\RETURN Selected model or not-recommended decision.
\end{algorithmic}
\end{algorithm}

\section{Experiments}
\label{subsec:operational_modes}

HantaWatch is designed as a user-facing federated surveillance framework rather than a single fixed classifier. This study evaluates the framework through two primary user modes. In Mode 1, training and model selection, users provide labelled historical Hantavirus sequence records. HantaWatch constructs task-specific datasets, configures federated clients, trains various federated methods, assesses surveillance metrics, and selects an appropriate model if deployment criteria are satisfied. In Mode 2, prediction-only triage, the selected model is applied to new unlabelled sequence records. The framework outputs predicted labels, risk scores, confidence values, uncertainty flags, review rationales, and expert-review priorities. A third mode, continuous update and monitoring, is planned for future implementation. In this mode, new labelled records would be incorporated regularly for retraining, concept drift monitoring, and ongoing surveillance.

\begin{figure}[htb]
\centering

\begin{subfigure}[t]{0.33\linewidth}
    \centering
    \includegraphics[width=\linewidth]{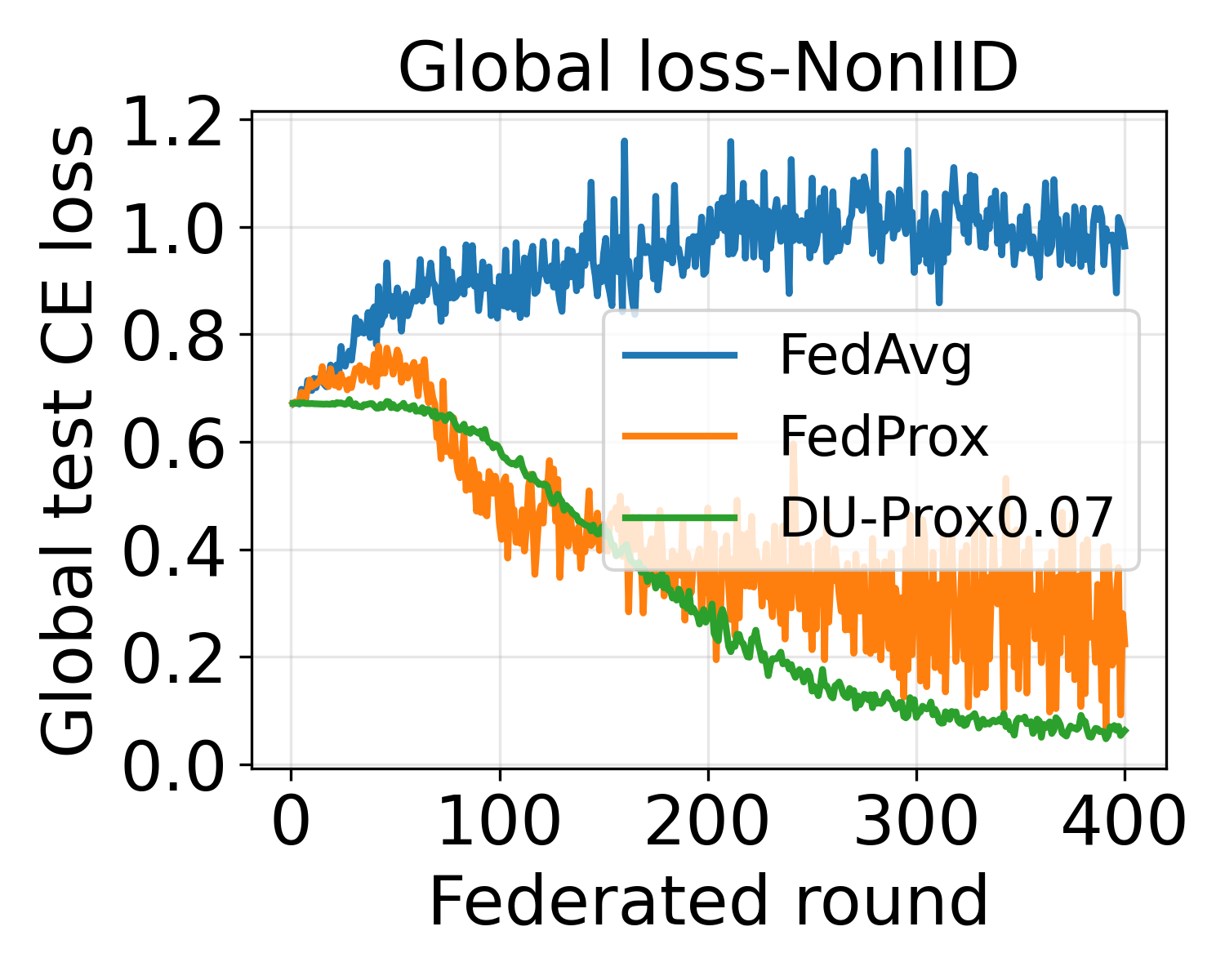}
    \caption{}
    \label{fig:HF_three_methods_global_ce_loss_noniid}
\end{subfigure}\hfill
\begin{subfigure}[t]{0.33\linewidth}
    \centering
    \includegraphics[width=\linewidth]{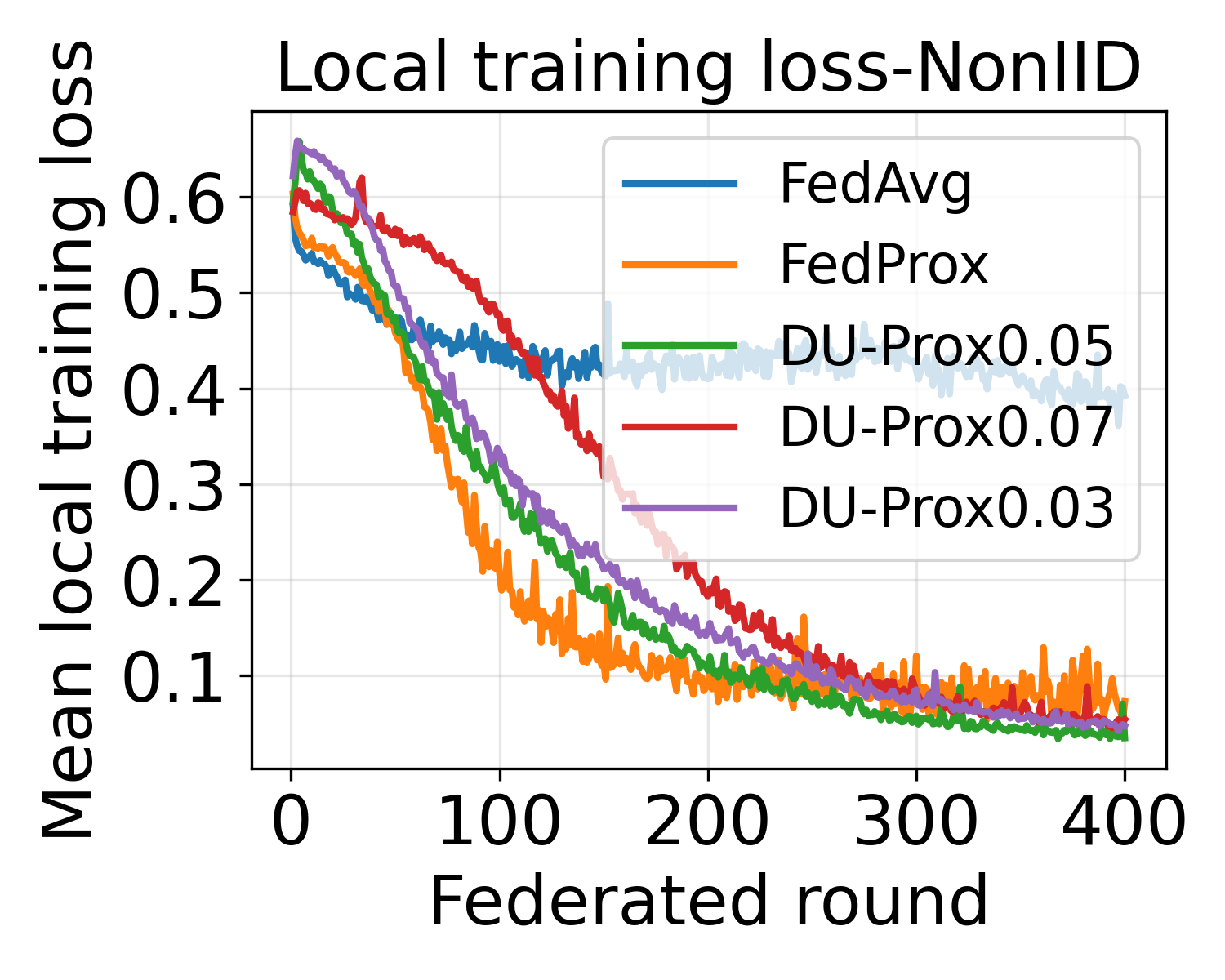}
    \caption{}
    \label{fig:HF_Local training loss over rounds noniid}
\end{subfigure}\hfill
\begin{subfigure}[t]{0.33\linewidth}
    \centering
    \includegraphics[width=\linewidth]{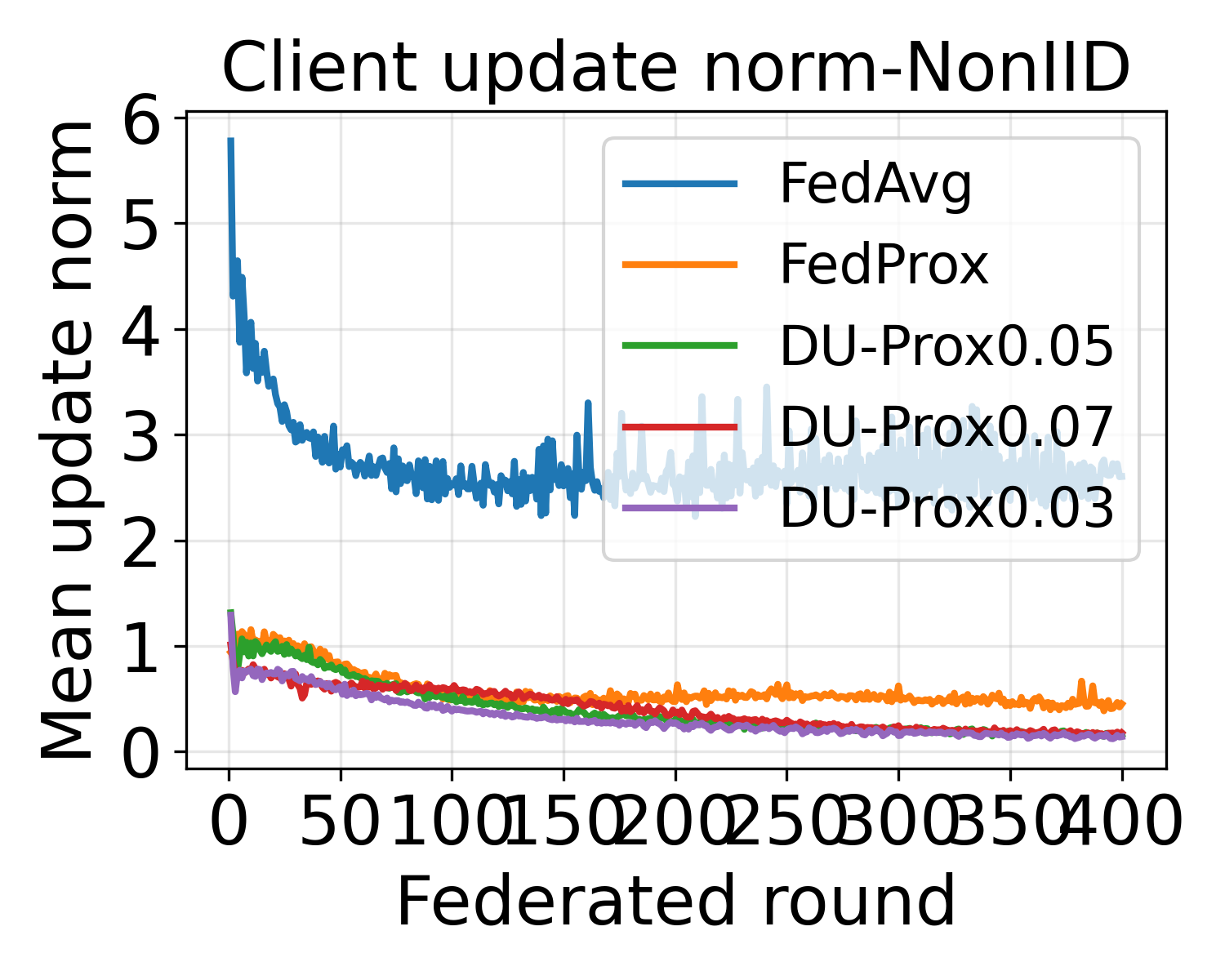}
    \caption{}
    \label{fig:HF_three_methods_update_norm_noniid}
\end{subfigure}\hfill

\begin{subfigure}[t]{0.33\linewidth}
    \centering
    \includegraphics[width=\linewidth]{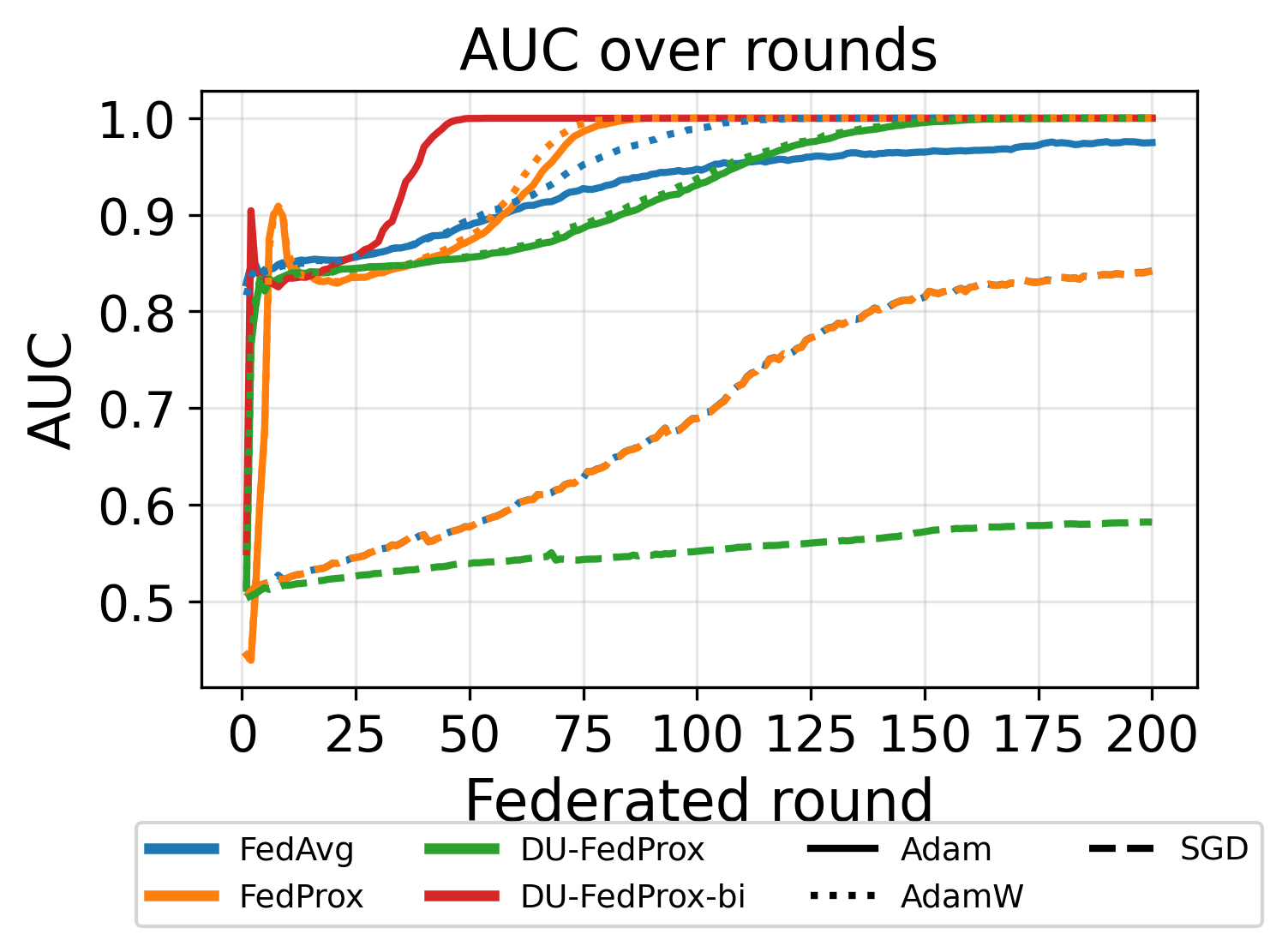}
    \caption{}
    \label{fig:optimizer_ablation_auc}
\end{subfigure}\hfill
\begin{subfigure}[t]{0.33\linewidth}
    \centering
    \includegraphics[width=\linewidth]{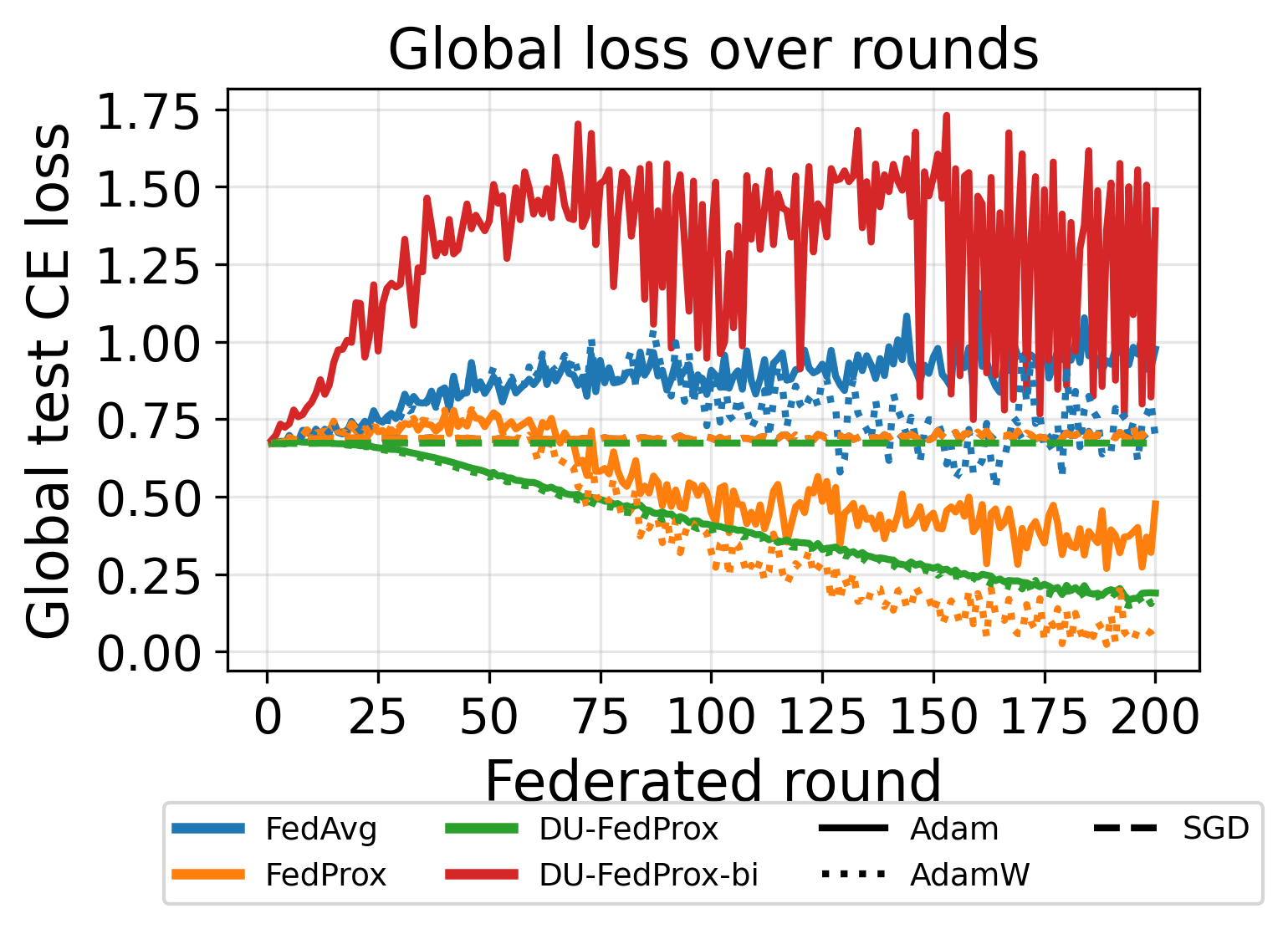}
    \caption{}
    \label{fig:optimizer_ablation_global_ce_loss}
\end{subfigure}\hfill
\begin{subfigure}[t]{0.33\linewidth}
    \centering
    \includegraphics[width=\linewidth]{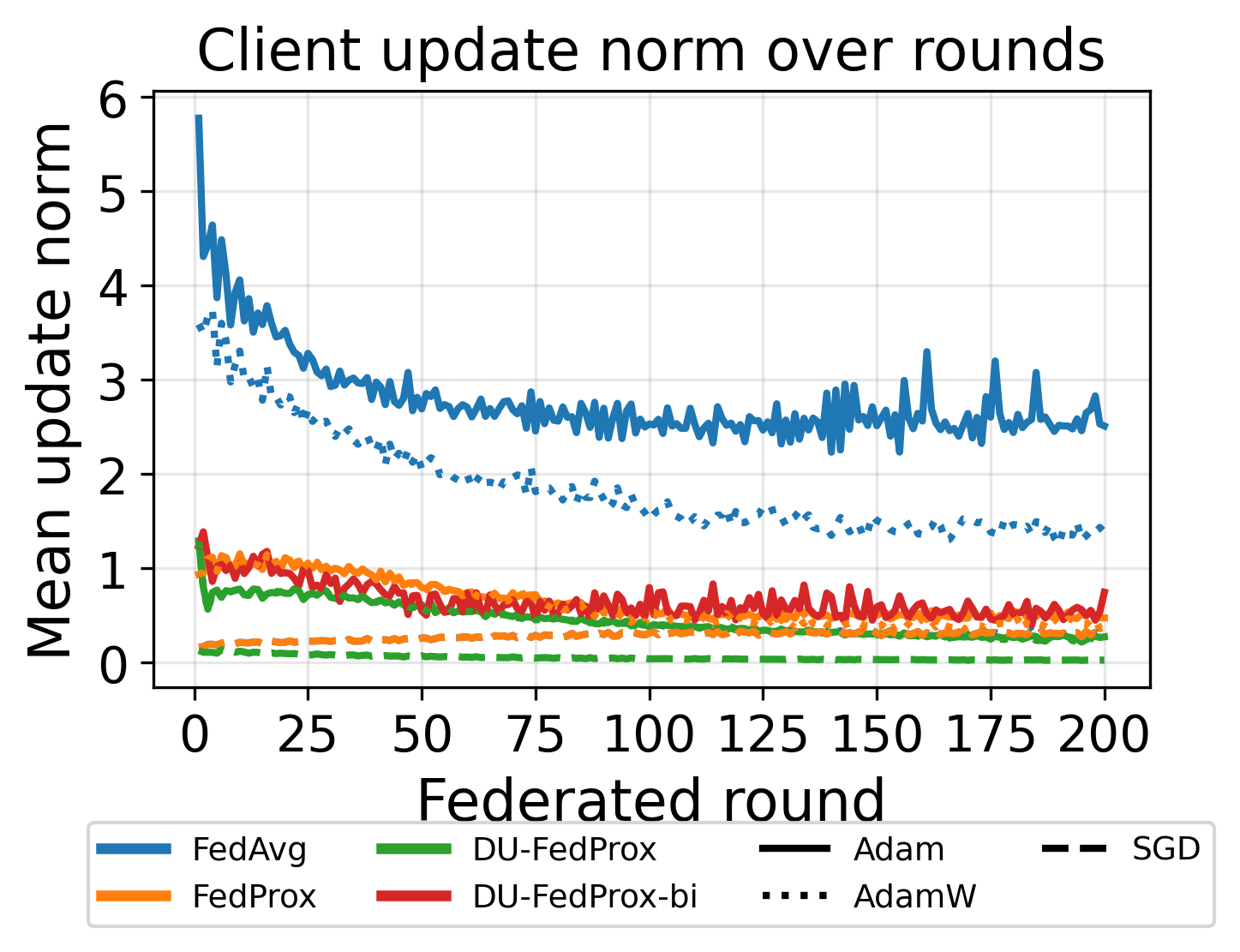}
    \caption{}
    \label{fig:optimizer_ablation_update_norm}
\end{subfigure}\hfill
\caption{HF high-pathogenicity results under different federated strategies and local optimizer settings. The plots show AUC, global loss, and client-update norm over federated rounds, demonstrating both predictive convergence and optimizer-dependent update stability.}
\label{fig:HFAndOpt}
\end{figure}

\subsection{Binary Surveillance Results}
\label{subsec:binary_results}

\begin{table*}[t]
\centering
\caption{Binary Hantavirus surveillance results over repeated non-IID federated runs. Values are reported as mean $\pm$ standard deviation over three runs. For binary surveillance tasks, recall and FNR are particularly important because missed surveillance-positive records may delay expert review. Higher accuracy, F1, recall, and AUC are preferred, while lower FNR and lower update norm are preferred.}
\label{tab:binary_results_summary}
\scriptsize
\renewcommand{\arraystretch}{1}
\begin{tabularx}{\textwidth}{
>{\raggedright\arraybackslash}p{2.5cm}
>{\raggedright\arraybackslash}p{2.2cm}
>{\centering\arraybackslash}p{1.55cm}
>{\centering\arraybackslash}p{1.55cm}
>{\centering\arraybackslash}p{1.55cm}
>{\centering\arraybackslash}p{1.55cm}
>{\centering\arraybackslash}p{1.55cm}
>{\centering\arraybackslash}p{1.55cm}
}
\toprule
\textbf{Task} &
\textbf{Method} &
\textbf{Accuracy} &
\textbf{F1} &
\textbf{Recall} &
\textbf{FNR} &
\textbf{AUC} &
\textbf{Update Norm} \\
\midrule

HF high-pathogenicity & FedAvg
& $0.633 \pm 0.004$
& $0.767 \pm 0.002$
& $\mathbf{1.000 \pm 0.000}$
& $\mathbf{0.000 \pm 0.000}$
& $0.994 \pm 0.000$
& $1.542 \pm 0.084$ \\

HF high-pathogenicity & FedProx
& $0.913 \pm 0.065$
& $0.935 \pm 0.046$
& $\mathbf{1.000 \pm 0.000}$
& $\mathbf{0.000 \pm 0.000}$
& $0.994 \pm 0.000$
& $0.349 \pm 0.048$ \\

HF high-pathogenicity & DU-FedProx-Single
& $\mathbf{0.960 \pm 0.004}$
& $\mathbf{0.968 \pm 0.003}$
& $\mathbf{1.000 \pm 0.000}$
& $\mathbf{0.000 \pm 0.000}$
& $\mathbf{0.995 \pm 0.000}$
& $\mathbf{0.160 \pm 0.028}$ \\

HF high-pathogenicity & DU-FedProx-Bilevel
& $0.663 \pm 0.008$
& $0.782 \pm 0.004$
& $\mathbf{1.000 \pm 0.000}$
& $\mathbf{0.000 \pm 0.000}$
& $0.994 \pm 0.000$
& $0.528 \pm 0.047$ \\

\midrule

HF outbreak-associated & FedAvg
& $\mathbf{0.916 \pm 0.008}$
& $\mathbf{0.913 \pm 0.008}$
& $\mathbf{0.933 \pm 0.013}$
& $\mathbf{0.067 \pm 0.013}$
& $\mathbf{0.961 \pm 0.005}$
& $0.961 \pm 0.020$ \\

HF outbreak-associated & FedProx
& $0.894 \pm 0.018$
& $0.892 \pm 0.016$
& $0.928 \pm 0.029$
& $0.072 \pm 0.029$
& $0.959 \pm 0.004$
& $0.351 \pm 0.031$ \\

HF outbreak-associated & DU-FedProx-Single
& $0.889 \pm 0.005$
& $0.879 \pm 0.007$
& $0.854 \pm 0.013$
& $0.146 \pm 0.013$
& $0.956 \pm 0.005$
& $\mathbf{0.062 \pm 0.002}$ \\

HF outbreak-associated & DU-FedProx-Bilevel
& $0.905 \pm 0.000$
& $0.899 \pm 0.000$
& $0.896 \pm 0.003$
& $0.104 \pm 0.003$
& $0.958 \pm 0.004$
& $0.487 \pm 0.031$ \\

\bottomrule
\end{tabularx}
\end{table*}

Table~\ref{tab:binary_results_summary} and Figures \ref{fig:HFAndOpt} demonstrate that HantaWatch elucidates trade-offs among candidate federated methods, contingent on the specific task. For high-pathogenicity screening, DU-FedProx-Single exhibited the best overall binary performance, achieving the highest accuracy, F1-score, AUC, and the lowest update norm, as well as perfect recall and zero FNR. In outbreak-associated screening, FedAvg achieved the highest predictive metrics, whereas DU-FedProx-Single again yielded the lowest update norm. These results indicate that adaptive proximal control can enhance update stability, although it does not consistently yield optimal performance across all tasks. Collectively, these findings validate HantaWatch's utility as a model-selection framework that systematically compares candidate methods and selects the most appropriate model according to the surveillance criteria relevant to each task.

For binary classification tasks, FedAvg and FedProx are established baselines. DU-FedProx extends these methods by adapting local optimization parameters, such as learning rate, proximal coefficient, and local epochs, according to each client's state. DU-FedProx demonstrates not only improved predictive performance but also enhanced training stability. In multiple binary scenarios, it reduces the client update norm and produces a smoother training loss curve, indicating superior control over client-update drift in the presence of client data heterogeneity. This improvement is particularly significant for surveillance applications, where unstable local updates can compromise the reliability of the global model.

Both predictive and surveillance-focused metrics should be considered when interpreting binary results. A method with high accuracy may still be suboptimal if it exhibits low recall or a high false negative rate, as this could result in missing samples critical for surveillance. Therefore, the binary analysis prioritizes F1-score, recall, area under the curve (AUC), and false negative rate (FNR), in addition to accuracy. If DU-FedProx achieves comparable accuracy while also reducing the update norm or false negative rate, it enhances the reliability of federated surveillance training.


\begin{table*}[t]
\centering
\caption{Multiclass Hantavirus surveillance results over repeated 200-round non-IID federated runs. Values are reported as mean $\pm$ standard deviation over three runs. For multiclass tasks, F1, recall, FNR, and AUC are macro-averaged. Lower macro-FNR and lower update norm are preferred.}
\label{tab:multiclass_results_200}
\scriptsize
\begin{tabular}{llcccccc}
\toprule
\textbf{Task} & \textbf{Method} & \textbf{Accuracy} & \textbf{Macro-F1} & \textbf{Macro-Recall} & \textbf{Macro-FNR} & \textbf{Macro-AUC} & \textbf{Update Norm} \\
\midrule
HF clade & FedAvg
& $0.919 \pm 0.025$
& $\mathbf{0.625 \pm 0.067}$
& $\mathbf{0.650 \pm 0.070}$
& $\mathbf{0.350 \pm 0.070}$
& --
& $1.401 \pm 0.033$ \\

HF clade & FedProx
& $0.910 \pm 0.008$
& $0.596 \pm 0.003$
& $0.622 \pm 0.017$
& $0.378 \pm 0.017$
& --
& $0.892 \pm 0.018$ \\

HF clade & DU-FedProx-Single
& $0.767 \pm 0.080$
& $0.356 \pm 0.026$
& $0.397 \pm 0.027$
& $0.603 \pm 0.027$
& --
& $\mathbf{0.469 \pm 0.023}$ \\

HF clade & DU-FedProx-Bilevel
& $\mathbf{0.925 \pm 0.009}$
& $0.582 \pm 0.013$
& $0.615 \pm 0.015$
& $0.385 \pm 0.015$
& --
& $0.969 \pm 0.041$ \\
\midrule
HF clinical syndrome & FedAvg
& $0.987 \pm 0.004$
& $0.986 \pm 0.005$
& $0.983 \pm 0.006$
& $0.017 \pm 0.006$
& $1.000 \pm 0.000$
& $1.139 \pm 0.090$ \\

HF clinical syndrome & FedProx
& $0.984 \pm 0.010$
& $0.982 \pm 0.014$
& $0.980 \pm 0.013$
& $0.020 \pm 0.013$
& $1.000 \pm 0.000$
& $0.618 \pm 0.061$ \\

HF clinical syndrome & DU-FedProx-Single
& $0.974 \pm 0.002$
& $0.972 \pm 0.002$
& $0.966 \pm 0.003$
& $0.034 \pm 0.003$
& $1.000 \pm 0.000$
& $\mathbf{0.455 \pm 0.094}$ \\

HF clinical syndrome & DU-FedProx-Bilevel
& $\mathbf{0.990 \pm 0.002}$
& $\mathbf{0.988 \pm 0.003}$
& $\mathbf{0.986 \pm 0.003}$
& $\mathbf{0.014 \pm 0.003}$
& $1.000 \pm 0.000$
& $0.614 \pm 0.044$ \\
\bottomrule
\end{tabular}
\end{table*}

\begin{table*}[h]
\centering
\caption{Comparison of user facing binary Hantavirus workflow results under two client-construction settings. The controlled label-skew split deliberately partitions one dataset into heterogeneous clients to evaluate robustness under non-IID conditions. The source-representative split constructs clients from source- or metadata-defined groups to approximate more realistic surveillance heterogeneity. Values are reported as mean $\pm$ standard deviation over three seeds.}
\label{tab:mode1_binary_training_comparison}
\scriptsize
\renewcommand{\arraystretch}{1.15}
\begin{tabular}{llcccccc}
\toprule
\textbf{Client construction} &
\textbf{Method} &
\textbf{Accuracy} &
\textbf{F1} &
\textbf{Precision} &
\textbf{Recall} &
\textbf{FNR} &
\textbf{Update Norm} \\
\midrule

Controlled label-skew split & FedAvg
& $0.990 \pm 0.000$
& $0.991 \pm 0.000$
& $0.986 \pm 0.005$
& $\mathbf{0.997 \pm 0.005}$
& $\mathbf{0.003 \pm 0.005}$
& $0.917 \pm 0.068$ \\

Controlled label-skew split & FedProx
& $0.803 \pm 0.171$
& $0.788 \pm 0.226$
& $0.925 \pm 0.026$
& $0.739 \pm 0.329$
& $0.261 \pm 0.329$
& $0.696 \pm 0.140$ \\

Controlled label-skew split & DU-FedProx-Single
& $0.948 \pm 0.023$
& $0.956 \pm 0.018$
& $0.966 \pm 0.037$
& $0.948 \pm 0.000$
& $0.052 \pm 0.000$
& $\mathbf{0.362 \pm 0.062}$ \\

Controlled label-skew split & DU-FedProx-Bilevel
& $\mathbf{0.990 \pm 0.005}$
& $\mathbf{0.991 \pm 0.004}$
& $\mathbf{0.986 \pm 0.010}$
& $\mathbf{0.997 \pm 0.005}$
& $\mathbf{0.003 \pm 0.005}$
& $0.461 \pm 0.018$ \\

\midrule

Source-representative split & FedAvg
& $0.870 \pm 0.077$
& $0.921 \pm 0.051$
& $\mathbf{0.996 \pm 0.002}$
& $0.859 \pm 0.086$
& $0.141 \pm 0.086$
& $2.113 \pm 0.301$ \\

Source-representative split & FedProx
& $0.855 \pm 0.079$
& $0.912 \pm 0.053$
& $0.989 \pm 0.004$
& $0.849 \pm 0.092$
& $0.151 \pm 0.092$
& $0.841 \pm 0.037$ \\

Source-representative split & DU-FedProx-Single
& $0.870 \pm 0.062$
& $0.923 \pm 0.040$
& $0.980 \pm 0.001$
& $0.873 \pm 0.070$
& $0.127 \pm 0.070$
& $\mathbf{0.505 \pm 0.048}$ \\

Source-representative split & DU-FedProx-Bilevel
& $\mathbf{0.930 \pm 0.055}$
& $\mathbf{0.960 \pm 0.034}$
& $0.984 \pm 0.007$
& $\mathbf{0.938 \pm 0.068}$
& $\mathbf{0.062 \pm 0.068}$
& $0.901 \pm 0.043$ \\

\bottomrule
\end{tabular}
\end{table*}

\begin{table*}[htb]
\centering
\caption{Comparison of user-facing multiclass clinical-syndrome training results under Mode 1 controlled label skew split and Mode 2 non-IID source representative split. Mode 1 divides the same dataset into artificially heterogeneous clients, while Mode 2 evaluates source/grouped-client heterogeneity. Values are reported as mean $\pm$ standard deviation over three seeds. For multiclass tasks, F1, recall, FNR, and AUC are macro-averaged.}
\label{tab:multiclass_mode1_mode2_training_comparison}
\scriptsize
\renewcommand{\arraystretch}{1.15}
\begin{tabular}{llcccccc}
\toprule
\textbf{Setting} &
\textbf{Method} &
\textbf{Accuracy} &
\textbf{Macro-F1} &
\textbf{Macro-Recall} &
\textbf{Macro-FNR} &
\textbf{Macro-AUC} &
\textbf{Update Norm} \\
\midrule

Controlled label-skew split& FedAvg
& $0.590 \pm 0.060$
& $0.441 \pm 0.080$
& $0.489 \pm 0.082$
& $0.511 \pm 0.082$
& $0.848 \pm 0.012$
& $1.410 \pm 0.117$ \\

Controlled label-skew split & FedProx
& $0.421 \pm 0.012$
& $0.164 \pm 0.025$
& $0.258 \pm 0.014$
& $0.742 \pm 0.014$
& $0.716 \pm 0.019$
& $\mathbf{0.761 \pm 0.032}$ \\

Controlled label-skew split & DU-FedProx-Single
& $0.414 \pm 0.000$
& $0.146 \pm 0.000$
& $0.250 \pm 0.000$
& $0.750 \pm 0.000$
& $0.763 \pm 0.042$
& $0.776 \pm 0.053$ \\

Controlled label-skew split & DU-FedProx-Bilevel
& $\mathbf{0.635 \pm 0.074}$
& $\mathbf{0.520 \pm 0.086}$
& $\mathbf{0.554 \pm 0.059}$
& $\mathbf{0.446 \pm 0.059}$
& $\mathbf{0.886 \pm 0.031}$
& $0.962 \pm 0.053$ \\

\midrule
Source-representative split& FedAvg
& $0.724 \pm 0.036$
& $0.730 \pm 0.043$
& $0.736 \pm 0.040$
& $0.264 \pm 0.040$
& $0.924 \pm 0.025$
& $1.876 \pm 0.221$ \\

Source-representative split& FedProx
& $0.452 \pm 0.048$
& $0.374 \pm 0.013$
& $0.433 \pm 0.016$
& $0.567 \pm 0.016$
& $0.782 \pm 0.004$
& $1.329 \pm 0.038$ \\

Source-representative split & DU-FedProx-Single
& $0.518 \pm 0.012$
& $0.473 \pm 0.031$
& $0.497 \pm 0.024$
& $0.503 \pm 0.024$
& $0.805 \pm 0.005$
& $\mathbf{0.932 \pm 0.050}$ \\

Source-representative split& DU-FedProx-Bilevel
& $\mathbf{0.734 \pm 0.062}$
& $\mathbf{0.739 \pm 0.075}$
& $\mathbf{0.752 \pm 0.067}$
& $\mathbf{0.248 \pm 0.067}$
& $\mathbf{0.932 \pm 0.031}$
& $0.826 \pm 0.011$ \\

\bottomrule
\end{tabular}
\end{table*}

\begin{figure}[htb]
\centering

\begin{subfigure}[t]{0.33\linewidth}
    \centering
    \includegraphics[width=\linewidth]{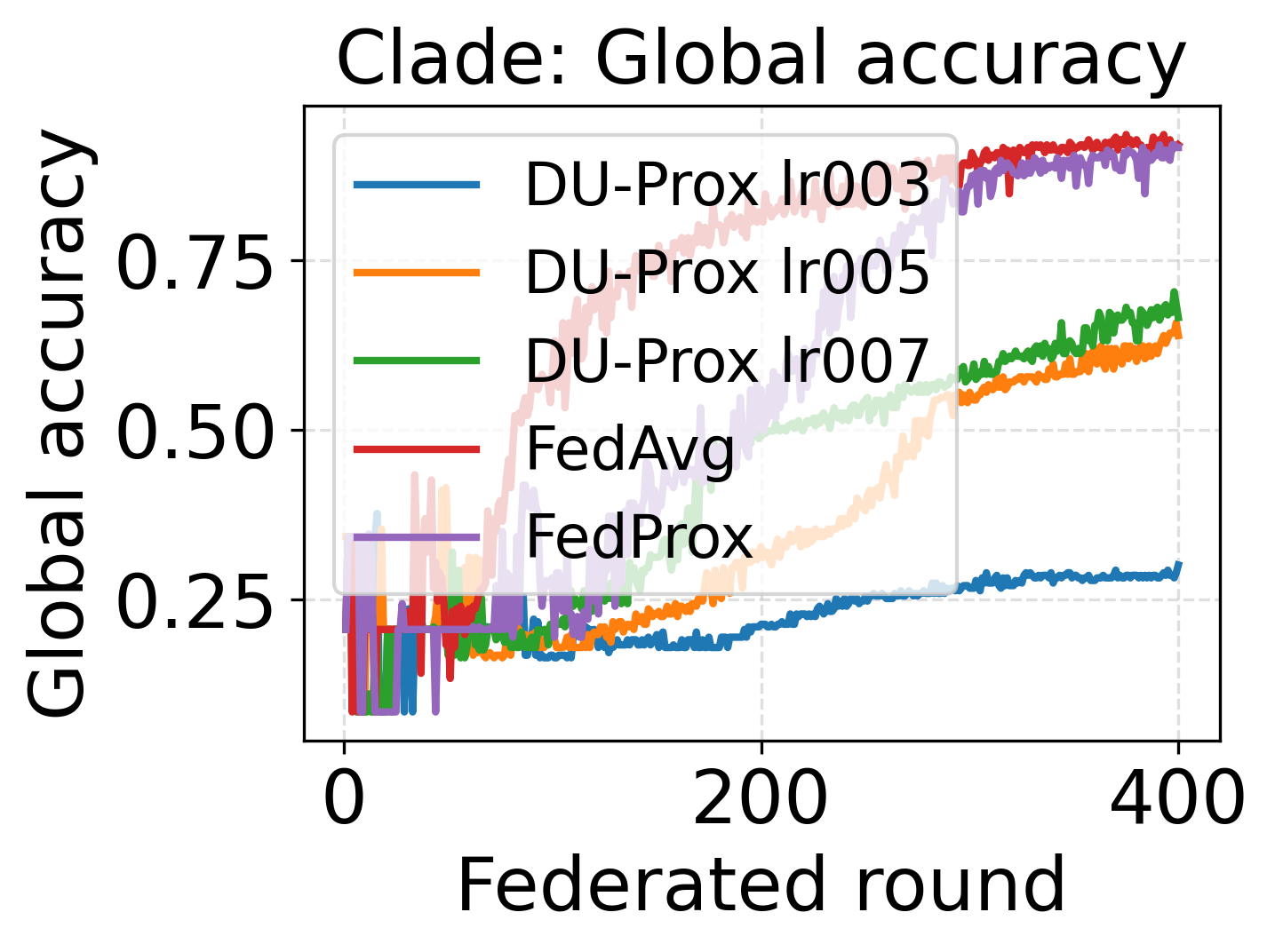}
    \caption{}
    \label{fig:ibm_duqfl_prox}
\end{subfigure}\hfill
\begin{subfigure}[t]{0.33\linewidth}
    \centering
    \includegraphics[width=\linewidth]{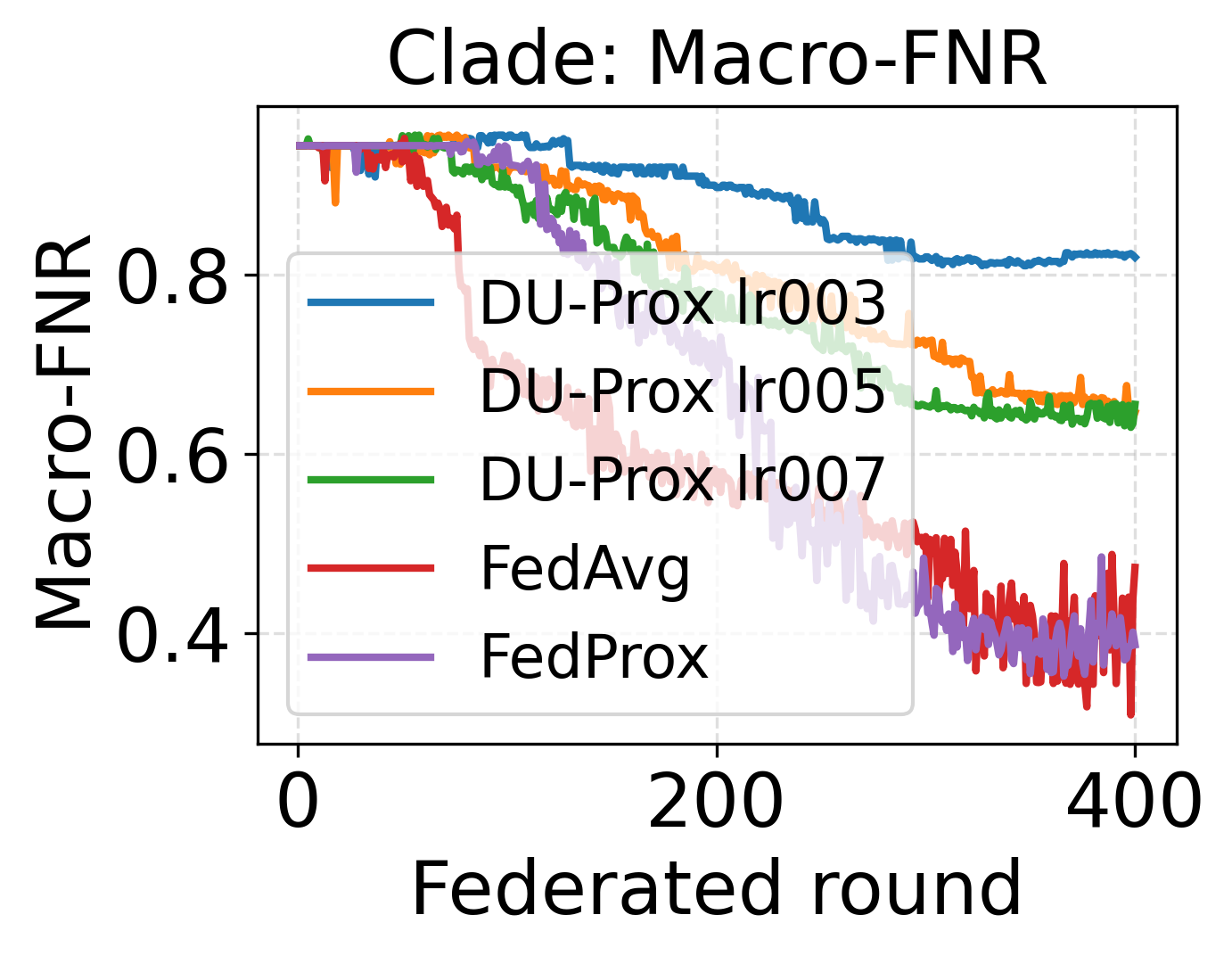}
    \caption{}
    \label{fig:CladeFNR}
\end{subfigure}\hfill
\begin{subfigure}[t]{0.33\linewidth}
    \centering
    \includegraphics[width=\linewidth]{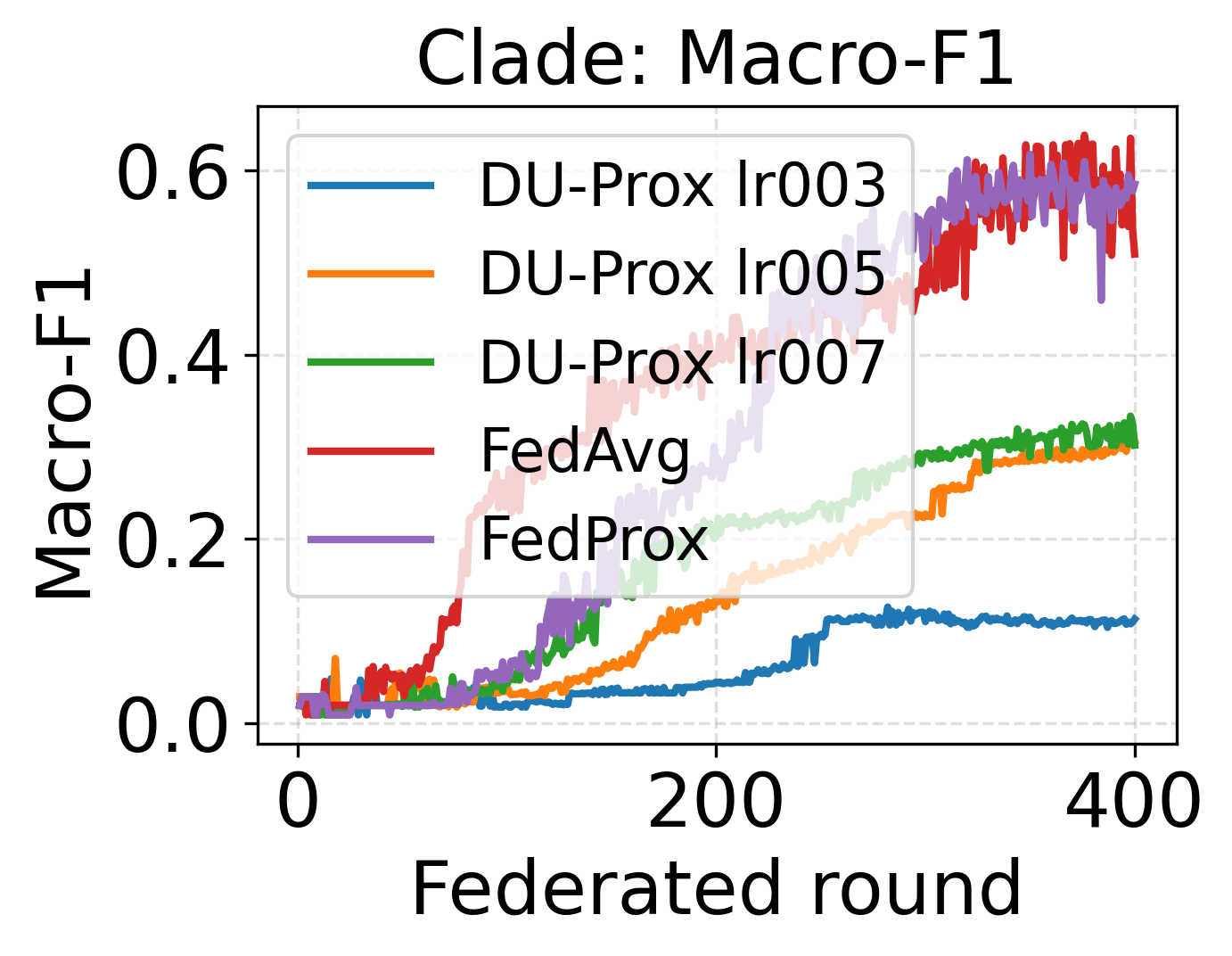}
    \caption{}
    \label{fig:ibm_default_qfl}
\end{subfigure}\hfill

\caption{Multiclass Clade classification}
\label{fig:Multiclass Clade classification}
\end{figure}

\begin{figure}[htb]
\centering

\begin{subfigure}[t]{0.33\linewidth}
    \centering
    \includegraphics[width=\linewidth]{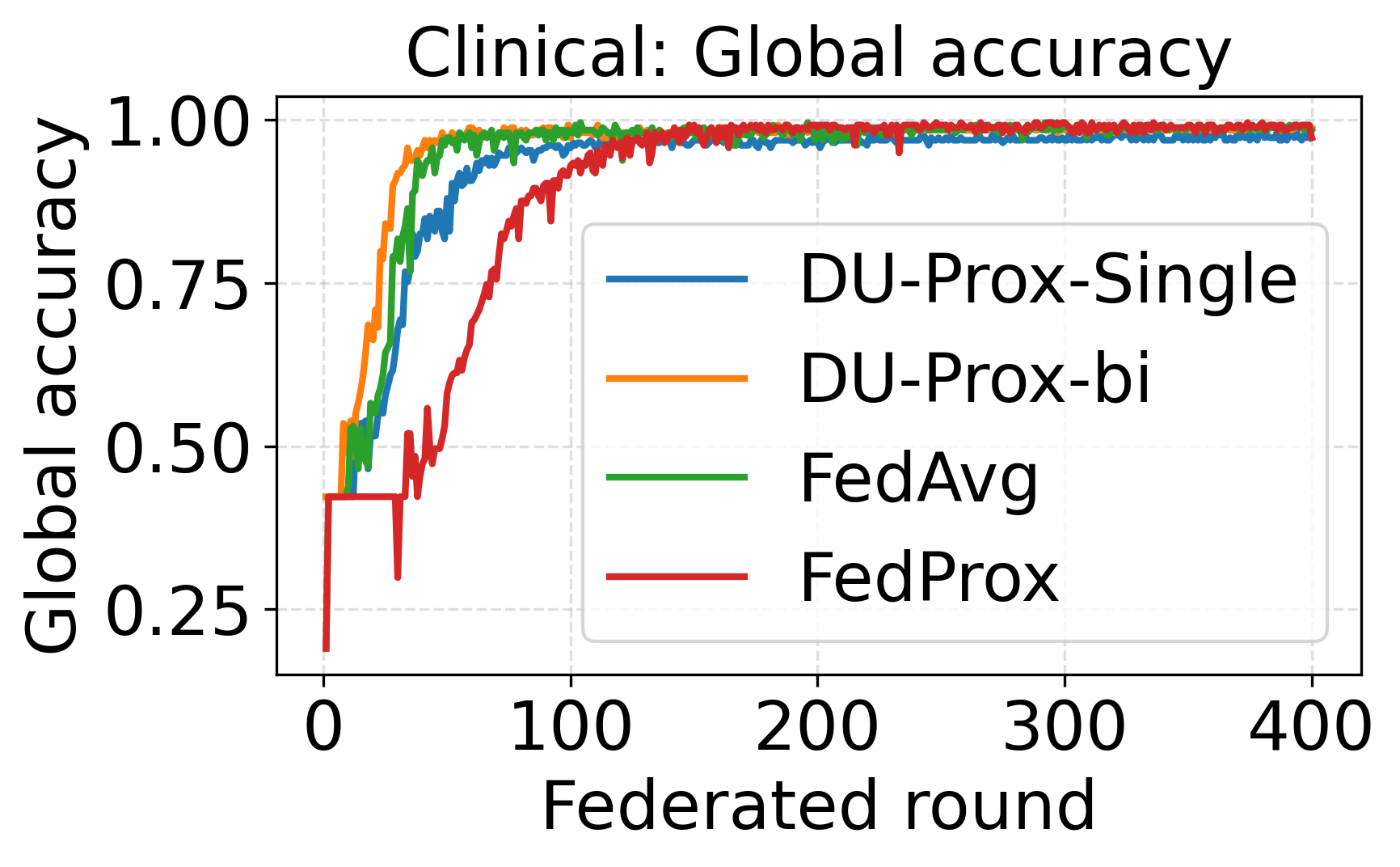}
    \caption{}
    \label{fig:clinical_syndrome_accuracy_curve}
\end{subfigure}\hfill
\begin{subfigure}[t]{0.33\linewidth}
    \centering
    \includegraphics[width=\linewidth]{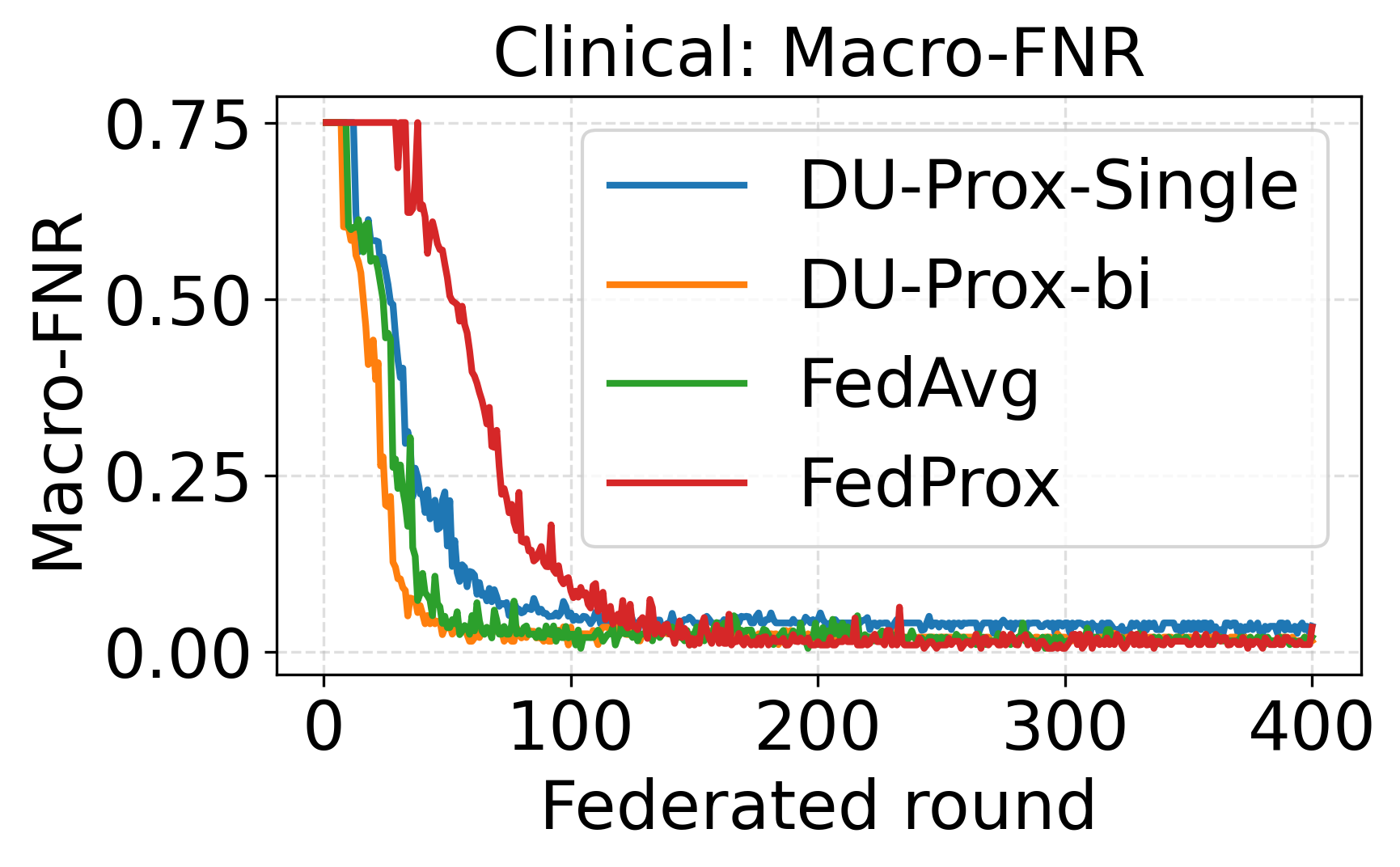}
    \caption{}
    \label{fig:CladeFNR}
\end{subfigure}\hfill
\begin{subfigure}[t]{0.33\linewidth}
    \centering
    \includegraphics[width=\linewidth]{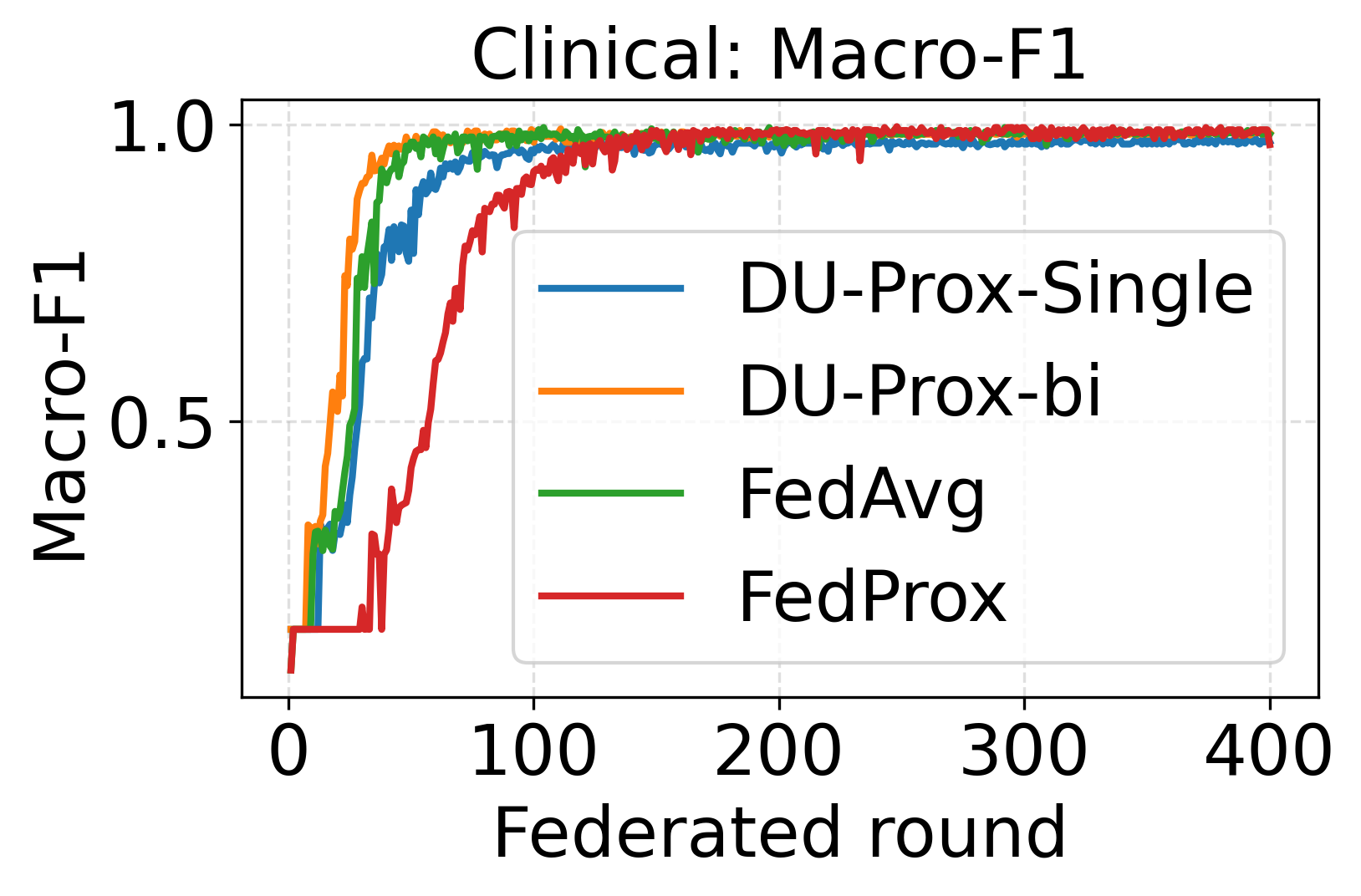}
    \caption{}
    \label{fig:ibm_default_qfl}
\end{subfigure}\hfill

\caption{Multiclass Clinical classification}
\label{fig:Multiclass Clinical classification}
\end{figure}

\begin{figure}[htb]
\centering

\begin{subfigure}[t]{0.33\linewidth}
    \centering
    \includegraphics[width=\linewidth]{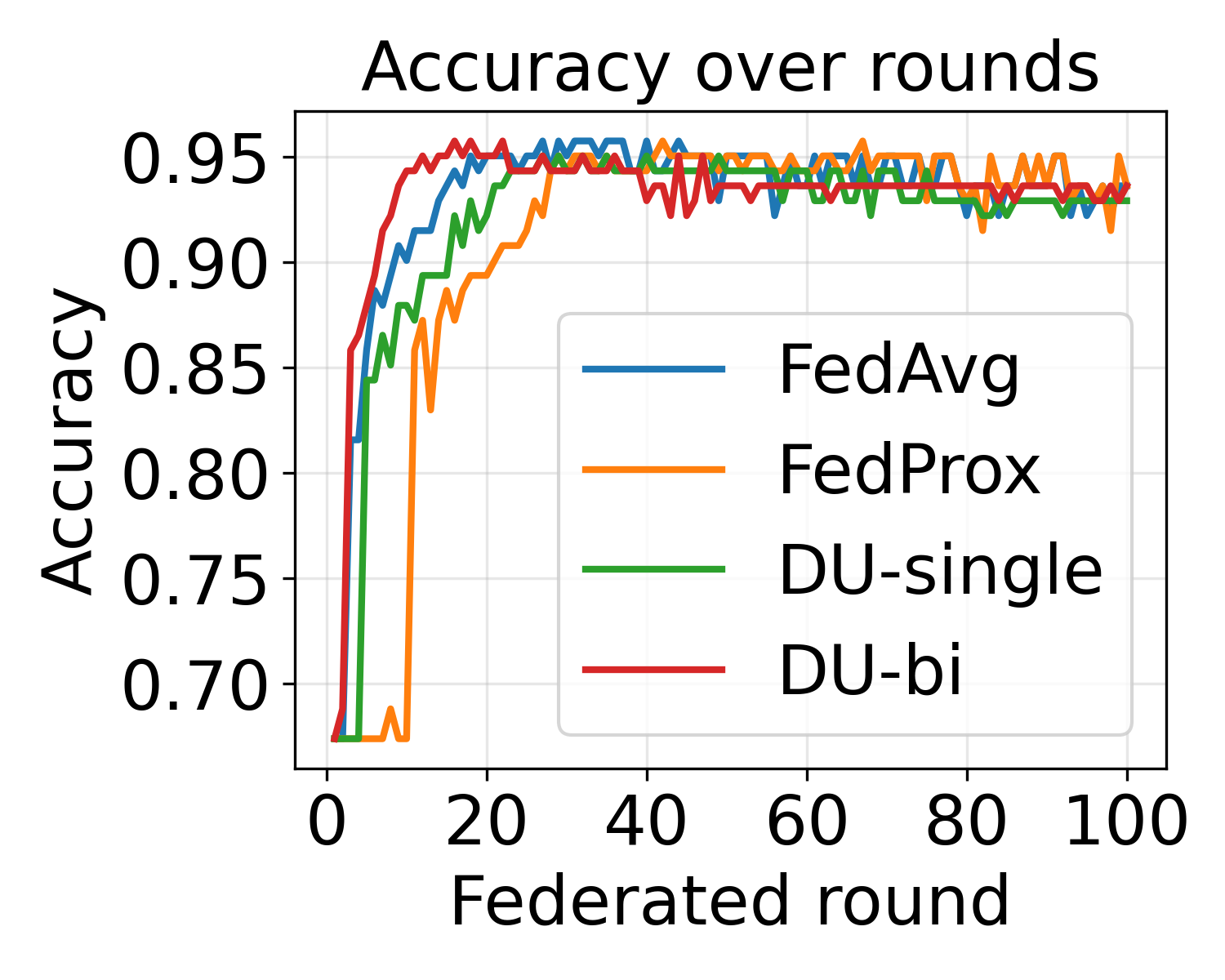}
    \caption{}
    \label{fig:ibm_duqfl_prox}
\end{subfigure}\hfill
\begin{subfigure}[t]{0.33\linewidth}
    \centering
    \includegraphics[width=\linewidth]{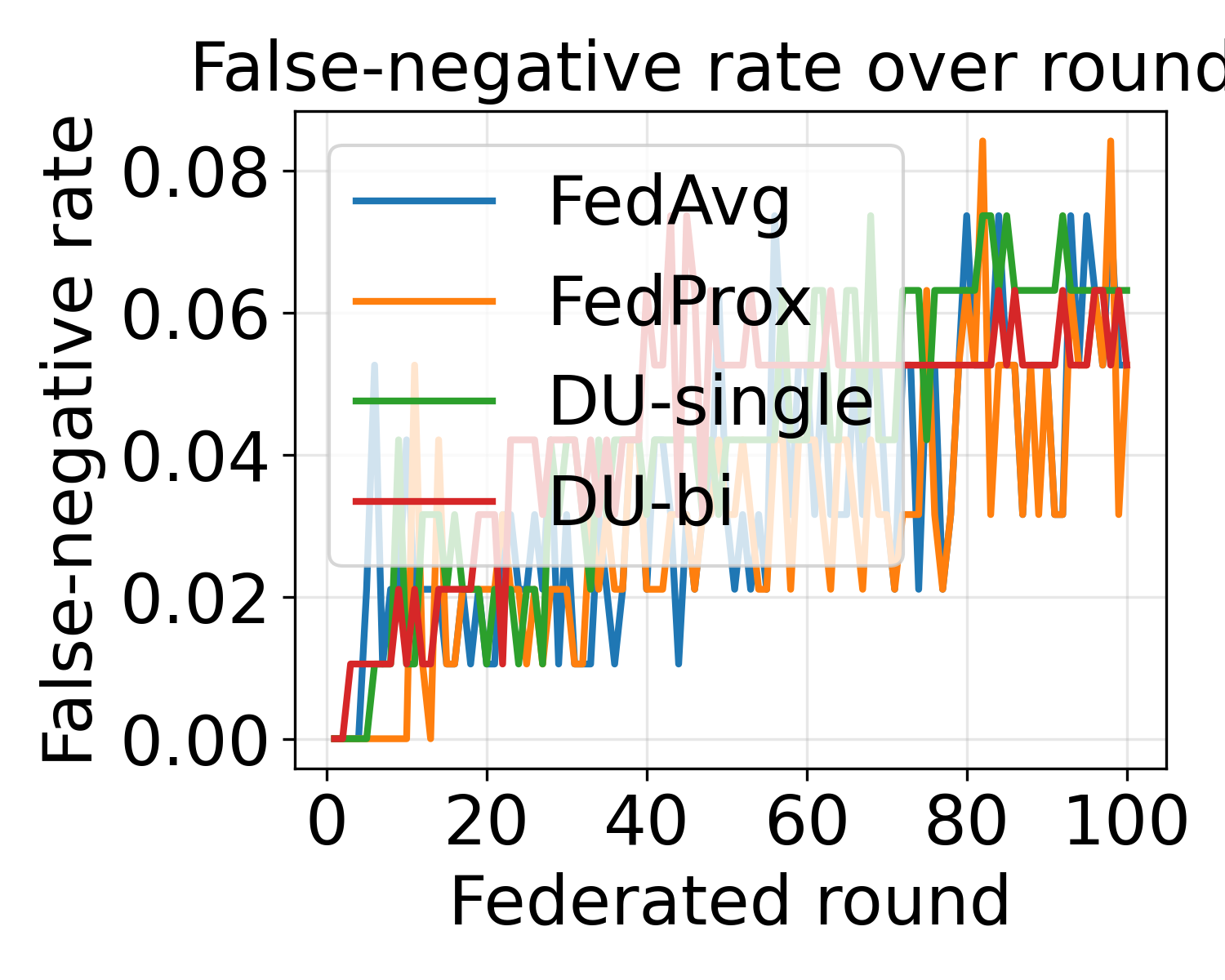}
    \caption{}
    \label{fig:three_methods_false_negative_rate_NCBI_nonIID}
\end{subfigure}\hfill
\begin{subfigure}[t]{0.33\linewidth}
    \centering
    \includegraphics[width=\linewidth]{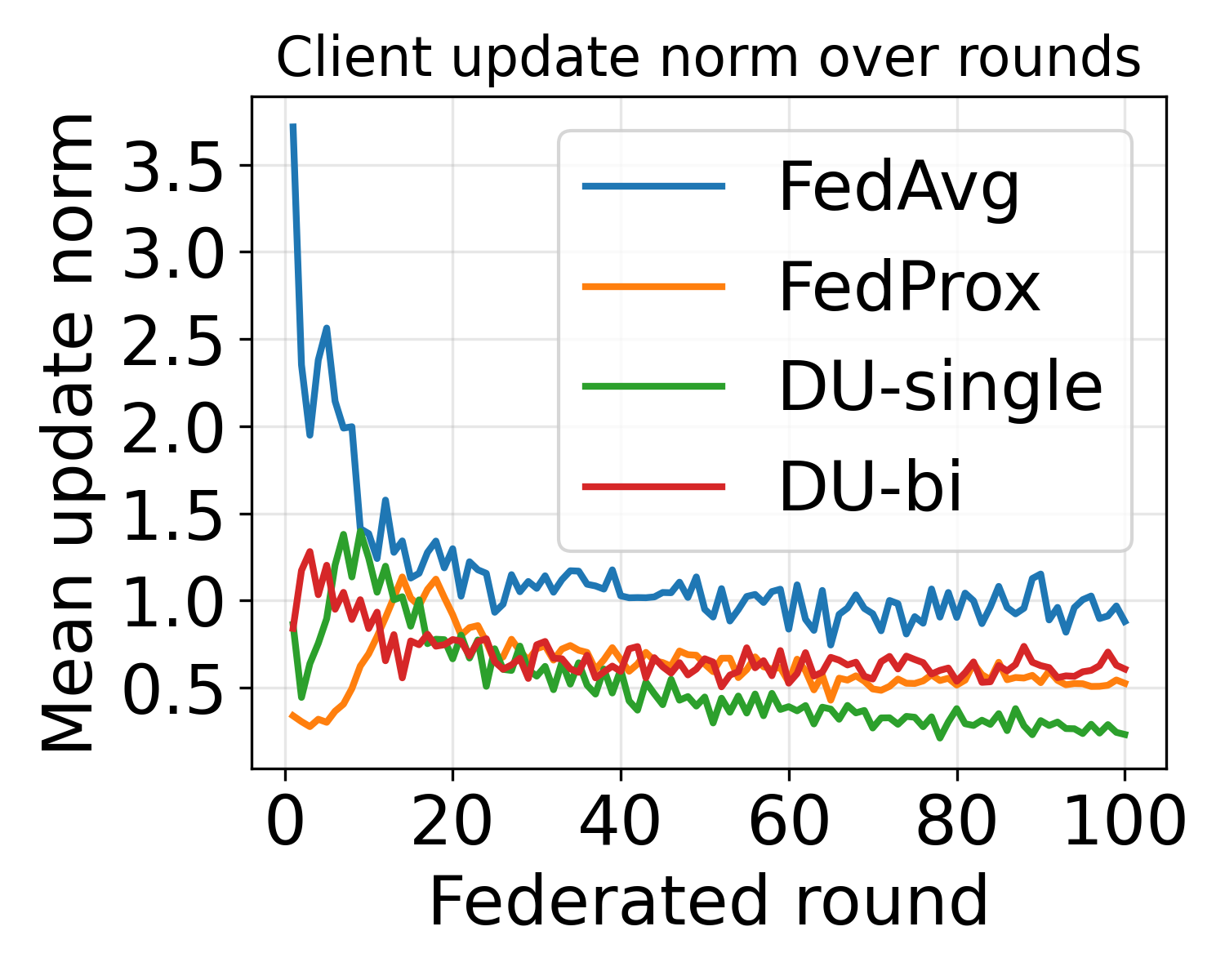}
    \caption{}
    \label{fig:three_methods_update_norm_NCBI_nonIID}
\end{subfigure}\hfill

\begin{subfigure}[t]{0.33\linewidth}
    \centering
    \includegraphics[width=\linewidth]{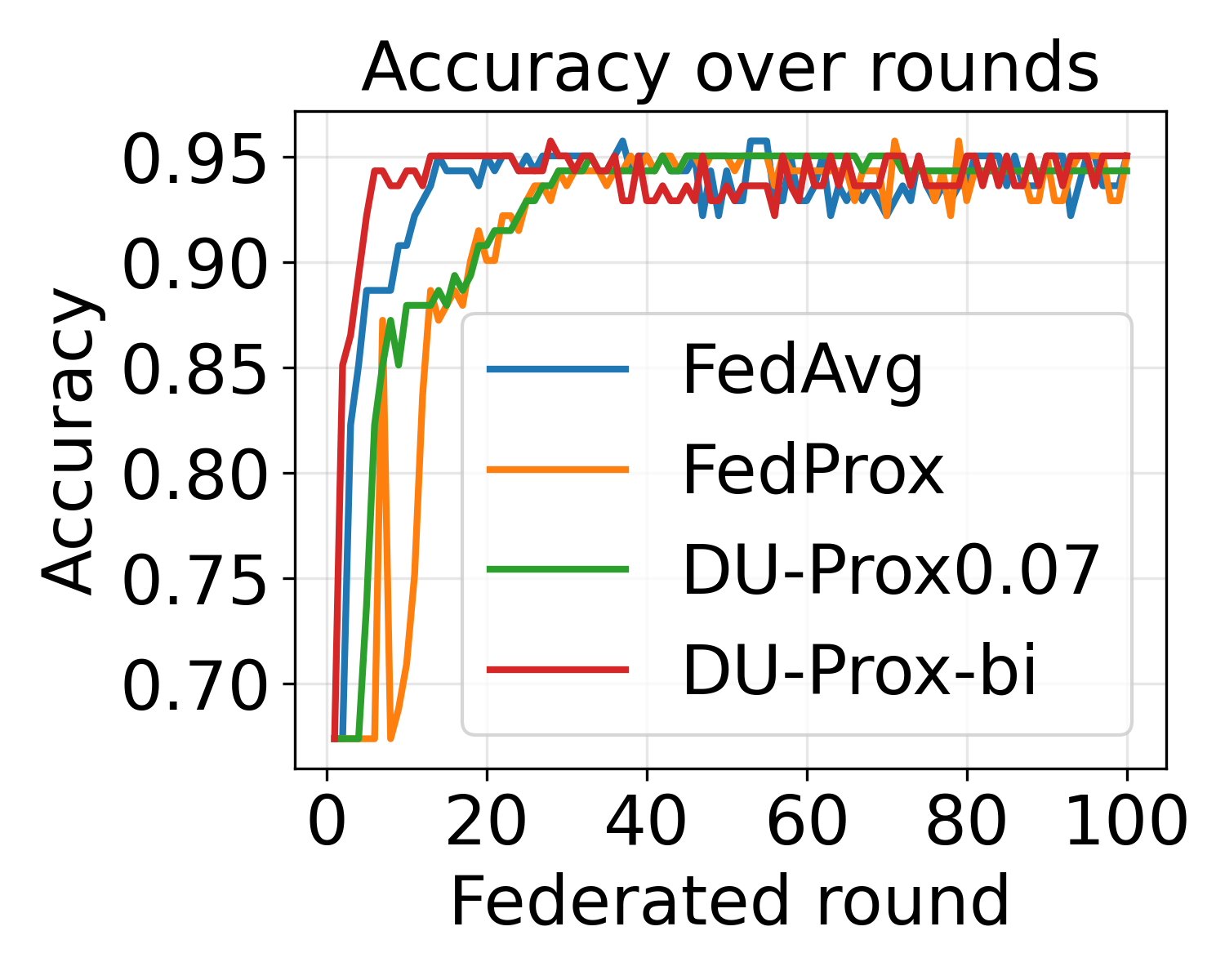}
    \caption{}
    \label{fig:three_methods_accuracy_NCBI_IID}
\end{subfigure}\hfill
\begin{subfigure}[t]{0.33\linewidth}
    \centering
    \includegraphics[width=\linewidth]{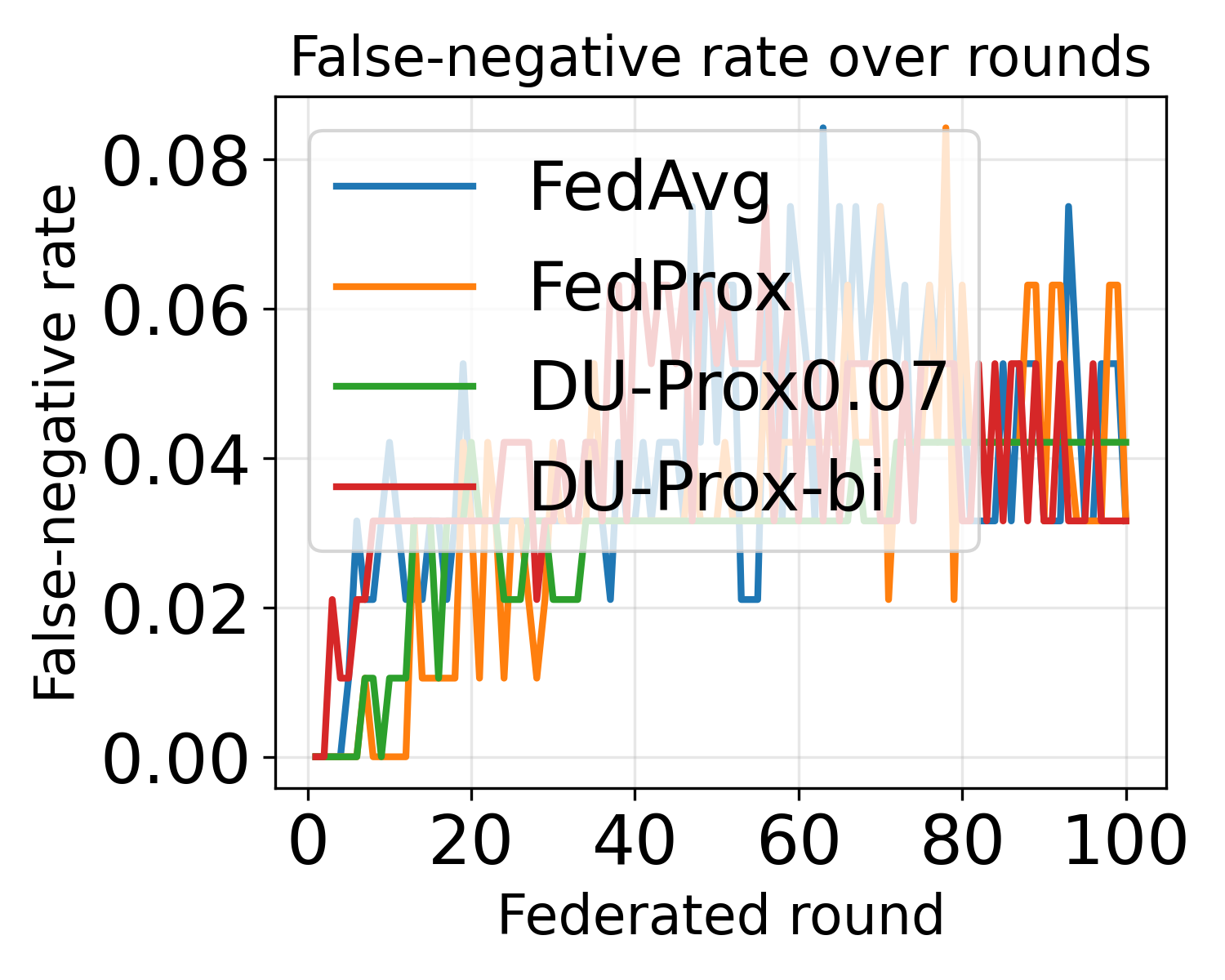}
    \caption{}
    \label{fig:three_methods_false_negative_rate_NCBI_IID}
\end{subfigure}\hfill
\begin{subfigure}[t]{0.33\linewidth}
    \centering
    \includegraphics[width=\linewidth]{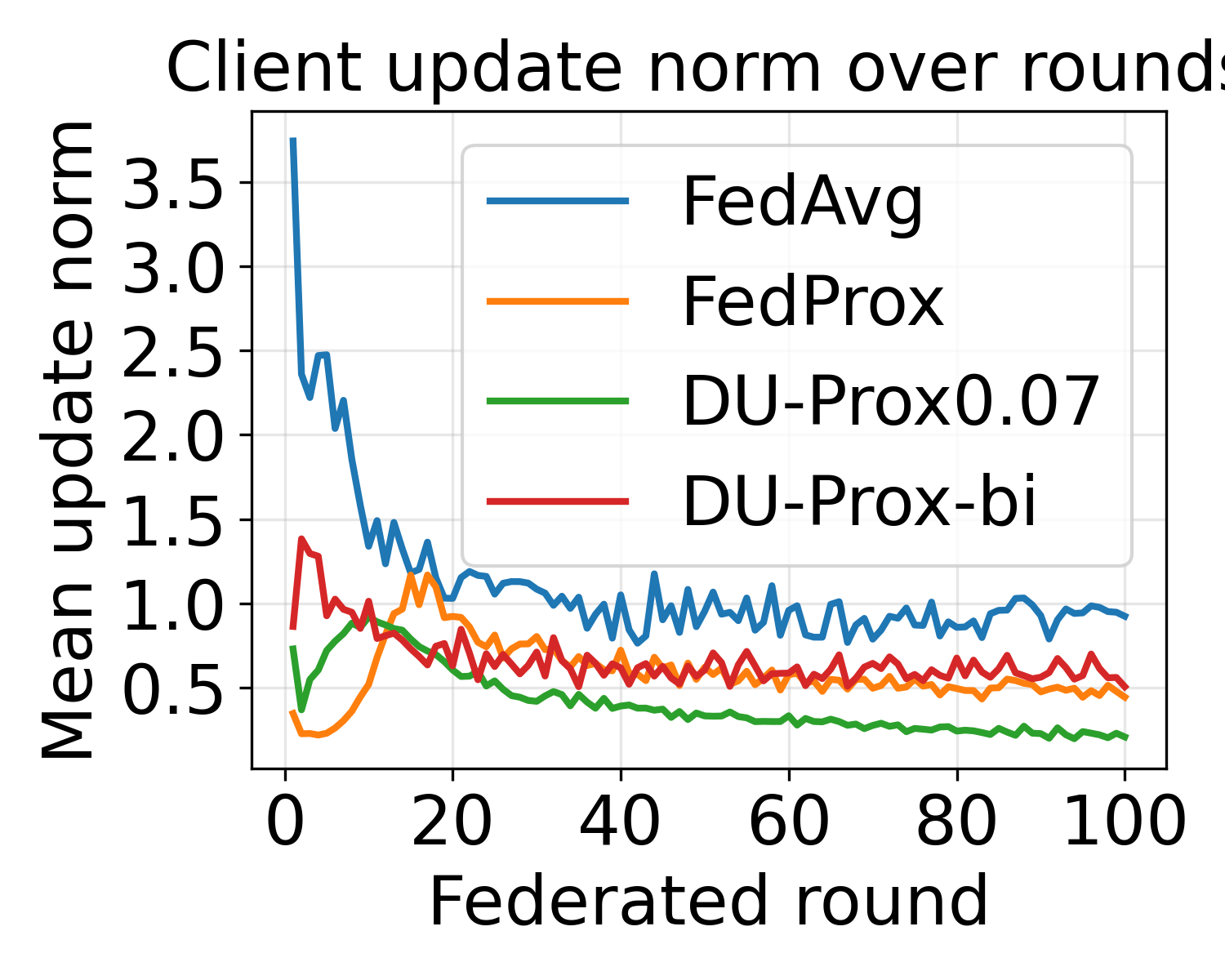}
    \caption{}
    \label{fig:three_methods_update_norm_NCBI_IID}
\end{subfigure}\hfill
\caption{NCBI federated convergence under non-IID and IID settings. The plots compare AUC, false-negative rate, and client update norm for FedAvg, FedProx, DU-FedProx-Single, and DU-FedProx-Bilevel.}
\label{fig:ncbi_convergence}
\end{figure}

\subsection{Multiclass Surveillance Results}
\label{subsec:Multiclass}

The clinical syndrome task evaluates HantaWatch within a four-class multiclass framework. The corresponding results are presented in Figure \ref{fig:Multiclass Clinical classification}.
Figure \ref{fig:Multiclass Clinical classification} indicates that all evaluated methods achieve strong performance in classifying clinical syndromes. These findings demonstrate that the framework is capable of handling both multiclass surveillance categorization and binary risk screening.
The results indicate that DU-FedProx-Bilevel achieves the highest overall performance, with a final accuracy of $0.9884$, macro-F1 of $0.9861$, macro-recall of $0.9846$, macro-AUC of $0.9999$, and macro-FNR of $0.0154$. These metrics suggest that the bilevel controller provides the most balanced surveillance performance among the evaluated methods. In comparison, FedAvg attains a final accuracy of $0.9845$ and macro-F1 of $0.9836$, while FedProx achieves a final accuracy of $0.9729$ and macro-F1 of $0.9661$. Although DU-FedProx-Single does not match the performance of the bilevel variant, it exhibits the smallest mean update norm, indicating more effective stabilization of client updates.

The results indicate that FedAvg and FedProx achieve higher peak performance during training. For example, FedProx attains a macro-F1 score of $0.9965$ at round 248, and FedAvg achieves an accuracy of $0.9961$ at round 104. In contrast, DU-FedProx-Bilevel demonstrates the best performance at the conclusion of training. These findings suggest that, although fixed baselines can achieve strong peak results on this clinical task, the bilevel controller provides greater stability and more balanced performance by the final round.


Figure \ref{fig:Multiclass Clade classification} demonstrates that the clade classification task is more challenging than the clinical syndrome task due to a higher number of classes and greater class imbalance. In this context, FedAvg and FedProx achieve superior final accuracy and macro-F1 scores compared to the current DU-FedProx configurations. These results indicate that multiclass Hantavirus classification with numerous classes requires more precise controller calibration than the simpler clinical syndrome classification.

These findings reveal a significant limitation: the effectiveness of adaptive proximal control is task-dependent. While such controllers can enhance training stability, they may become overly conservative in challenging multiclass scenarios by excessively reducing the learning rate or over-regularizing local training. The clade task indicates that future research should focus on designing controllers tailored for multiclass problems, incorporating class-balanced objectives, and establishing task-specific proximal constraints.


\subsection{Optimizer and Training Stability}
\label{subsec:stability_results}

Figure~\ref{fig:HFAndOpt} illustrates the impact of local optimizer selection on candidate federated strategies. This analysis is significant because a user-facing surveillance framework should not depend on a single optimizer configuration that may only perform well for a particular dataset or client distribution. Experimental results indicate that optimizer choice influences convergence speed, AUC progression, loss trajectories, and the magnitude of client updates. Certain optimizer and method combinations resulted in more rapid predictive improvement but produced larger or more variable updates. In contrast, other combinations achieved smoother convergence and maintained more stable updates.

Figure~\ref{fig:ncbi_convergence} presents a comparison of FedAvg, FedProx, DU-FedProx-Single, and DU-FedProx-Bilevel in both non-IID and IID NCBI federated settings. The AUC curves initially increase rapidly before stabilizing. The FNR curves illustrate the risk of failing to detect positive cases. The update-norm curves indicate that DU-FedProx variants generally reduce the magnitude of client updates relative to standard FedAvg, suggesting improved control of client drift during heterogeneous training.


\subsection{User-Facing Workflow Evaluation}
\label{subsec:user_facing_workflow}

The user-facing evaluation assesses whether HantaWatch can execute the complete train, select, and predict workflow, rather than solely reporting classification metrics. In a simulated deployment, the framework selected DU-FedProx-Bilevel as the model and applied it to 300 pseudo-unlabelled records. The resulting priority list identified all 72 true high-priority records for expert review, achieving a triage recall of $1.000$, a review precision of $1.000$, and a review burden of $0.240$. The top-$k$ analysis indicated that the highest-ranked records were predominantly high-priority cases, with perfect precision among the top 10, 20, and 50 reviewed records. These findings demonstrate that HantaWatch can effectively translate model predictions into a functional expert-review queue.

The framework was evaluated on external records from NCBI using a prediction-only approach. In this assessment, HantaWatch processed 702 unlabelled sequences and generated risk scores, predicted labels, confidence values, uncertainty flags, and outputs for expert review. Due to the absence of curated, task-specific labels for this dataset, the primary objective was to assess the system's performance in real-world scenarios rather than to quantify prediction accuracy. The results indicate that HantaWatch is capable of processing external sequence records and producing structured review outputs for expert evaluation.

The source-controlled Mode 2 experiment provides a more realistic evaluation of client performance by grouping data to form clients, rather than relying solely on artificial label-skewed partitions. Under these conditions, DU-FedProx-Bilevel outperformed DU-FedProx-Single and resulted in a deployable model. These findings indicate that HantaWatch is applicable in scenarios extending beyond a single simulated non-independent and identically distributed (non-IID) partition. However, prediction-only deployment should be reported exclusively when the selected model path is consistently utilized during inference.

\subsection{Real-Data Evaluation on HuggingFace Hantavirus Records}
\label{subsec:realdata_results}

HantaWatch was evaluated using Hantavirus sequence records from the HuggingFace dataset. The framework was tested with features derived from the sequences and surveillance labels for specific tasks. Both binary and multiclass classification tasks were included in the evaluation.

Two approaches were used to construct federated clients. The first, termed the controlled label-skew setting, involves splitting a single labeled dataset into multiple clients to establish a reproducible non-IID benchmark. The second, referred to as the source-representative setting, groups clients according to source or metadata to mirror the variations observed among actual surveillance data holders. The controlled setting evaluates algorithm performance under known imbalances, whereas the source-representative setting assesses performance in the presence of more realistic, source-level differences.

Table~\ref{tab:mode1_binary_training_comparison} presents a comparison of the binary Hantavirus workflow under both experimental settings. Under controlled label-skew partitioning, the classification task was readily separable. FedAvg and DU-FedProx-Bilevel achieved near-perfect predictive performance. DU-FedProx-Single exhibited the lowest update norm, while DU-FedProx-Bilevel provided the most balanced performance for surveillance, as measured by accuracy, F1-score, recall, and false negative rate (FNR).

The source-representative setting presented greater challenges. DU-FedProx-Bilevel achieved the highest accuracy, F1-score, recall, and the lowest false negative rate (FNR). In contrast, FedAvg attained the highest area under the curve (AUC), although its update norm was substantially larger. These findings indicate that constructing clients at the source level accentuates methodological differences more than artificial label-skew partitioning. This approach also provides a more rigorous evaluation of the reliability of the federated surveillance.

Table~\ref{tab:multiclass_mode1_mode2_user_output} presents the prediction results for users following the Mode 1 workflow. In the artificial non-IID workflow, HantaWatch processed 524 records and flagged 367 for expert review, of which 363 were classified as high-risk and 4 were flagged due to uncertainty. In the source-as-client workflow, 420 records were processed and 380 were flagged for expert review, including 369 high-risk and 11 uncertainty-review records. The increased number of reviews in the source-as-client setting is attributed to a more diverse prediction scenario. These findings demonstrate that HantaWatch provides both evidence of model performance and clear review-priority lists for expert assessment.

\textit{Multiclass Clinical-Syndrome Classification}.
\label{subsubsec:real_multiclass_clinical} The clinical-syndrome experiment evaluates HantaWatch in a context involving multiple surveillance categories. The objective is not to automate clinical diagnoses, but rather to determine whether models developed from sequence data can support surveillance categorization and the prioritization of expert reviews. As in the clade-classification task, macro-averaged metrics are emphasized to prevent overestimation of performance in the presence of class imbalance.

\begin{table}[t]
\centering
\caption{Prediction-only user-output comparison under the controlled label-skew and source-representative workflows. The table summarizes the priority-review files generated after applying the selected workflow model to new or held-out sequence records.}
\label{tab:mode1_binary_user_output_comparison_transposed}
\scriptsize
\renewcommand{\arraystretch}{1.15}
\begin{tabular}{lcc}
\toprule
\textbf{Output item} &
\textbf{Controlled label-skew split} &
\textbf{Source-representative split} \\
\midrule
Records processed & 524 & 420 \\
Predicted positive & 369 & 383 \\
Predicted negative & 155 & 37 \\
High-risk review & 363 & 369 \\
Uncertainty review & 4 & 11 \\
Routine & 157 & 40 \\
Expert-review flagged & 367 & 380 \\
Review burden & 0.700 & 0.905 \\
Mean confidence & 0.984 & 0.948 \\
\bottomrule
\end{tabular}
\end{table}

\begin{table}[t]
\centering
\caption{Prediction-only user-output comparison for the clinical-syndrome multiclass workflow. The controlled label-skew split represents prediction output after Mode 1 training and model selection, while the source-representative split represents the Mode 2 output on held-out records.}
\label{tab:multiclass_mode1_mode2_user_output}
\scriptsize
\renewcommand{\arraystretch}{1.15}
\begin{tabular}{lcc}
\toprule
\textbf{Output item} &
\textbf{Controlled label-skew split} &
\textbf{Source-representative split} \\
\midrule
Records processed & 524 & 258 \\
Predicted HFRS & 263 & 76 \\
Predicted HPS & 17 & 40 \\
Predicted mild & 32 & 44 \\
Predicted non-human & 212 & 98 \\
Uncertainty review & 360 & 46 \\
Routine priority & 164 & 212 \\
Expert-review flagged & 360 & 46 \\
Review burden & 0.687 & 0.178 \\
Mean confidence & 0.563 & 0.821 \\
\bottomrule
\end{tabular}
\end{table}

\section{Conclusion}
\label{sec:conclusion}

We propose HantaWatch, a federated surveillance-support framework for Hantavirus sequence analysis and expert-review prioritization. The framework integrates sequence-derived feature extraction, non-IID and source-aware client construction, federated model comparison, adaptive DU-FedProx control, suitability assessment, and prediction-only review output generation. The results show that HantaWatch can support both binary surveillance screening and multiclass categorization while emphasizing surveillance-oriented metrics such as recall, false-negative rate, F1-score, AUC, and update stability. DU-FedProx improves update stability and provides performance benefits in selected tasks, although FedAvg and FedProx remain strong baselines in other settings, particularly for more difficult high-cardinality classification. Overall, HantaWatch should be interpreted as a decision-support and expert-prioritization system rather than an autonomous diagnostic tool, with future work focusing on stronger multiclass controller calibration, concept-drift monitoring, out-of-distribution detection, secure aggregation, privacy-preserving communication, and broader validation across multi-source and multi-pathogen surveillance settings.

\bibliographystyle{IEEEtran}
\bibliography{bib}

\end{document}